%% file: main.tex
\documentclass{article} 
\usepackage{iclr2027_conference,times}

\input{math_commands.tex}

\usepackage{hyperref}
\usepackage{url}
\usepackage{xcolor}%
\usepackage{textcomp}%
\usepackage{manyfoot}%
\usepackage{booktabs}%
\usepackage{algpseudocode}%
\usepackage{listings}%
\usepackage{balance}
\usepackage{mathtools}
\usepackage{tablefootnote} 
\usepackage{multirow} 
\usepackage{graphicx}
\usepackage{subfigure}
\usepackage[ruled,vlined]{algorithm2e}
\SetKwInput{KwIn}{Input}
\SetKwInput{KwOut}{Output}
\usepackage{adjustbox}
\usepackage{pifont}
\usepackage{amsmath,amssymb,amsfonts}%
\usepackage{amsthm}%
\usepackage{mathrsfs}%
\usepackage{float}
\usepackage[table]{xcolor}
\usepackage{cancel}
\usepackage{tcolorbox}
\usepackage{makecell}
\usepackage{caption}
\usepackage[dvipsnames]{xcolor}  

\title{OmniRSI: Taming Latent Diffusion for Unified Multimodal Remote Sensing Image Generation}

\author{Hebaixu Wang, Jing Zhang\footnotemark[1] , Haonan Guo, Di Wang, Jiayi Ma\footnotemark[1] , and Bo Du\thanks{Corresponding Authors.}  \\
	Wuhan University \& Zhongguancun Academy\\ 
	\texttt{\{wanghebaixu, jingzhang.cv, jyma2010\}@gmail.com} \\ 
	\texttt{\{haonan.guo, d\_wang, dubo\}@whu.edu.cn} 
}

\iclrfinalcopy 
\begin{document}
\maketitle 

\begin{abstract}
Multimodal remote sensing image generation emerges as a promising paradigm for recovering complete and complementary observations from heterogeneous and partially observed remote sensing data. However, existing methods predominantly depend on sensor-specific fine-tuning of natural image generation models, limiting unified multimodal generation capacity due to the absence of shared representations and robust conditioning mechanisms. To address these challenges, we propose OmniRSI, a unified multimodal remote sensing image generation model. Specifically, we develop a remote sensing-specific variational autoencoder that captures multimodal priors and promotes a locally smooth latent space for unified representation and robust reconstruction across diverse observed data. Building upon that, we fully adapt Stable Diffusion 3.5 with a ControlNet branch to support unified multimodal generation and translation conditioned on textual and visual cues. Besides, we devise a training-free cyclic latent alignment mechanism that iteratively enriches incomplete visual conditions with generated multimodal information, progressively improving guidance reliability while preserving observed measurements. We also construct RSI-MM, a large-scale dataset comprising 11.5M multi-resolution images with text annotations and partially paired data for training and evaluation. Extensive experiments verify that OmniRSI achieves superior performance in multimodal reconstruction, generation and translation. 
\end{abstract} 

\section{Introduction}\label{sec:intro} 
Multimodal remote sensing images are fundamental carriers for earth observation, where different modalities capture distinct yet complementary information of earth surface. Contemporary satellites are typically equipped with modality-specific sensors that acquire diverse observations, including panchromatic (PAN), multispectral (MSI), and synthetic aperture radar (SAR) imagery. These heterogeneous observations support a wide range of applications~\cite{Wu2025}. However, due to sensor-specific acquisition mechanisms, orbital constraints, and varying observation conditions, complete and spatially aligned multimodal observations are often unavailable in practice. This scarcity of paired multimodal data has motivated the development of generative models~\cite{shao2025generative,yu2026metaearth} for remote sensing image synthesis and completion.
 
Generative models~\cite{hu2026simulating} aim to learn underlying data distributions and synthesize novel samples with similar characteristics. Recent advances in diffusion models have pushed the frontiers of visual generation tasks due to their remarkable performance~\cite{he2025diffusion,xiao2025omnigen,esser2024scaling}. Most multimodal remote sensing image generation approaches follow the latent diffusion-based generative pipeline~\cite{zhao2026anyband,tang2025aerogen,yu2024metaearth}. It generally consists of three essential stages: representation, generation, and translation. At the first stage, some methods~\cite{zhang2025cc} directly utilize frozen variational autoencoders (VAEs) pretrained on natural images~\cite{rombach2022high}, which provide efficient latent compression but yield suboptimal representations for remote sensing data. Other methods~\cite{chen2026any2any} explore more compatible solutions by performing modality-specific VAE adaptation. At the second stage, most methods perform generative sampling in the latent space by gradually perturbing data through the forward diffusion process and learning a reverse denoising process to recover the underlying data distribution~\cite{fuest2026diffusion}. They typically adapt large-scale pretrained diffusion models via parameter-efficient fine-tuning to efficiently transfer general visual priors to remote sensing domains while retaining their inherent text-to-image (T2I) generation capabilities. However, these methods are usually developed for individual modalities, such as text-to-RGB~\cite{liu2025text2earth,zhang2024cc} and text-to-SAR generation~\cite{liu2026huiyanearth,fan2025smart}. At the third stage, auxiliary modality conditions are incorporated into generative models to guide the sampling process. These conditions are encoded and injected into the generation backbone through dedicated branches or attention-based mechanisms~\cite{ye2026object,zhang2026any2rsi}. Such extensions rely on predefined paired relationships, and can thus be viewed as task-specific visual translation, where a conditional distribution is learned from a source modality to a target modality.
 
\begin{figure*}[t]
	\centering
	\includegraphics[width=0.99\linewidth]{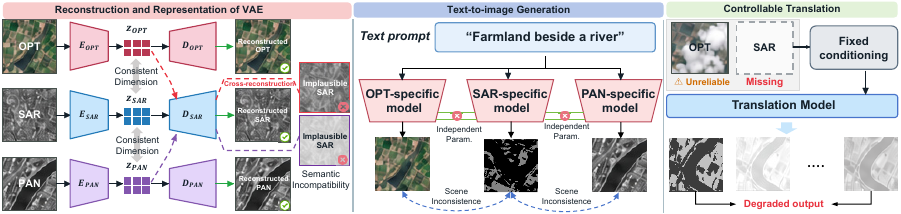}
	\vspace{-0.5em}
	\caption{Overview and limitations of latent diffusion-based remote sensing image generation.}
	\label{fig:notion}
\end{figure*}

Despite these advances, existing methods remain insufficient for multimodal remote sensing generation, exhibiting three fundamental limitations: semantically misaligned latent spaces, fragmented multimodal generation, and rigid conditional guidance (Fig.~\ref{fig:notion}). First, modality-specific VAE adaptation primarily ensures dimensional consistency while leaving substantial semantic discrepancies across modalities unresolved and limiting the representation transferability. Second, current generative models are typically adapted to individual modalities, preventing effective sharing of generative priors across heterogeneous observations and resulting in fragmented generation capabilities. Third, conditional translation methods rely on predefined source-target relationships and fixed guidance throughout inference, limiting the flexibility across modality combinations. When source observations are incomplete or uncertain, such static conditioning cannot adaptively refine unreliable guidance, making the generation process vulnerable to accumulated errors. More fundamentally, these components are largely developed in isolation, lacking an integrated design that jointly couples multimodal representation learning, unified generation, and adaptive condition refinement.

To address these challenges, we propose OmniRSI, a unified multimodal remote sensing generation model that tightly couples the representation, generation and translation. Specifically, we organize heterogeneous observations into modality-aware slots according to sensor-specific characteristics and develop a remote sensing-specific variational autoencoder (RS-VAE) to learn shared latent representations from incomplete and heterogeneous observations. By combining masked reconstruction with latent smoothness regularizations, it is able to capture cross-modal dependencies and promotes a locally smooth latent space, thereby enabling robust representation under incomplete observations. Building upon this, we adapt Stable Diffusion 3.5 with a ControlNet branch to achieve unified multimodal generation from textual descriptions and conditional translation from partially available observations. To improve the condition reliability and alleviate accumulated errors, we introduce a training-free cyclic latent alignment (CLA) mechanism that iteratively enriches incomplete visual conditions with generated multimodal information while preserving observed measurements during inference. To support training and evaluation, we construct RSI-MM, a large-scale multimodal remote sensing dataset containing 11.5 million multi-resolution global images with text annotations and partially paired observations. Extensive experiments demonstrate that OmniRSI consistently outperforms the state-of-the-art methods across reconstruction, generation, and translation tasks. 

Our main contributions can be summarized as follows: 1) We develop RS-VAE that learns a unified latent representation from incomplete and heterogeneous observations by leveraging modality-aware slots and tailored reconstruction strategies. 2) We adapt Stable Diffusion 3.5 with a ControlNet branch to tightly couple multimodal generation and translation. A training-free sampling mechanism is devised to progressively refine unreliable conditions while preserving observed measurements. 3) We construct RSI-MM, a large-scale multimodal remote sensing dataset with diverse global observations, text descriptions, and partially paired modalities for training and evaluation. 

\section{Related Work}\label{sec:related_work}
\textbf{Generative Models} can be primarily represented by autoregressive (AR) and diffusion-based approaches. Autoregressive models learn image distributions by predicting discrete visual tokens produced by pretrained tokenizers. Recent advances have explored both improved token representations and flexible prediction paradigms. For the former, RQ-VAE~\cite{lee2022autoregressive} introduces recursive residual quantization to reduce mapping errors, FSQ~\cite{mentzer2024finite} quantizes continuous features, and TokenFlow~\cite{qu2025tokenflow} explores multiple codebook for visual representation at different granularities. For the latter, recent methods~\cite{liu2024customize,wang2025parallelized,pang2026next} relax the fixed raster-order prediction paradigm through flexible token grouping, parallel prediction, and patch-level token aggregation, while VAR~\cite{tian2024visual} further introduces next-scale prediction for coarse-to-fine image synthesis. Diffusion models generate samples by progressively perturbing data and learning the reverse inference process. Latent Diffusion~\cite{rombach2022high} improves scalability by performing diffusion in a compressed latent space, enabling large-scale generative systems such as Stable Diffusion (SD) series~\cite{esser2024scaling,podell2024sdxl}. Beyond text conditioning, recent studies have introduced more flexible control mechanisms. Training-free methods, including classifier guidance~\cite{dhariwal2021diffusion}, classifier-free guidance~\cite{Gao_2026_CVPR,lin2026freecontrol}, and inversion-based editing methods~\cite{xie2026dnaedit}, manipulate sampling trajectories without additional training. Meanwhile, fine-tuning methods~\cite{hu2025sara} such as T2I-Adapter~\cite{mou2024t2i} and ControlNet~\cite{zhang2023adding,motorcu2026pg} incorporate external conditions by lightweight adapters or additional control branches.

\textbf{Remote sensing image generation} typical follows diffusion-based paradigms and can be categorized into text- and visual-conditioned generation. For text-conditioned generation, several studies investigate modality-specific remote sensing image synthesis. Txt2Img-MHN~\cite{xu2023txt2img} utilizes hierarchical prototype learning to improve scene consistency for RGB generation, while Text2Earth~\cite{liu2025text2earth} constructs a large-scale dataset and develops a model for generating diverse optical scenes. HSIGene~\cite{pang2026hsigene} introduces spectral-aware generation strategies and multi-condition controllability for hyperspectral image generation. HuiYanEarth-SAR~\cite{liu2026huiyanearth} explores text-guided SAR image generation by incorporating SAR-specific imaging characteristics and specific priors. Beyond textual conditions, recent studies introduce additional visual or structural guidance to improve controllability and spatial consistency. Any2RSI~\cite{zhang2026any2rsi} explores controllable generation by incorporating diverse control signals and enriched descriptions. DSP~\cite{bao2026envisioning} investigates layout-to-image generation under limited data regimes by disentangling semantic concepts and visual primitives. CC-Diff~\cite{zhang2024cc}, AeroGen~\cite{tang2025aerogen}, and OFDiff~\cite{ye2026object} incorporate geographic or observation-specific conditions to improve the fidelity and enable region-aware controllability. However, these approaches generally focus on modality-specific generation with predefined conditioning, leaving unified generation across heterogeneous modalities from partially observed multimodal inputs still unexplored. 

\section{Method}\label{sec:method}
\begin{figure*}[h]
	\centering
	\includegraphics[width=0.99\linewidth]{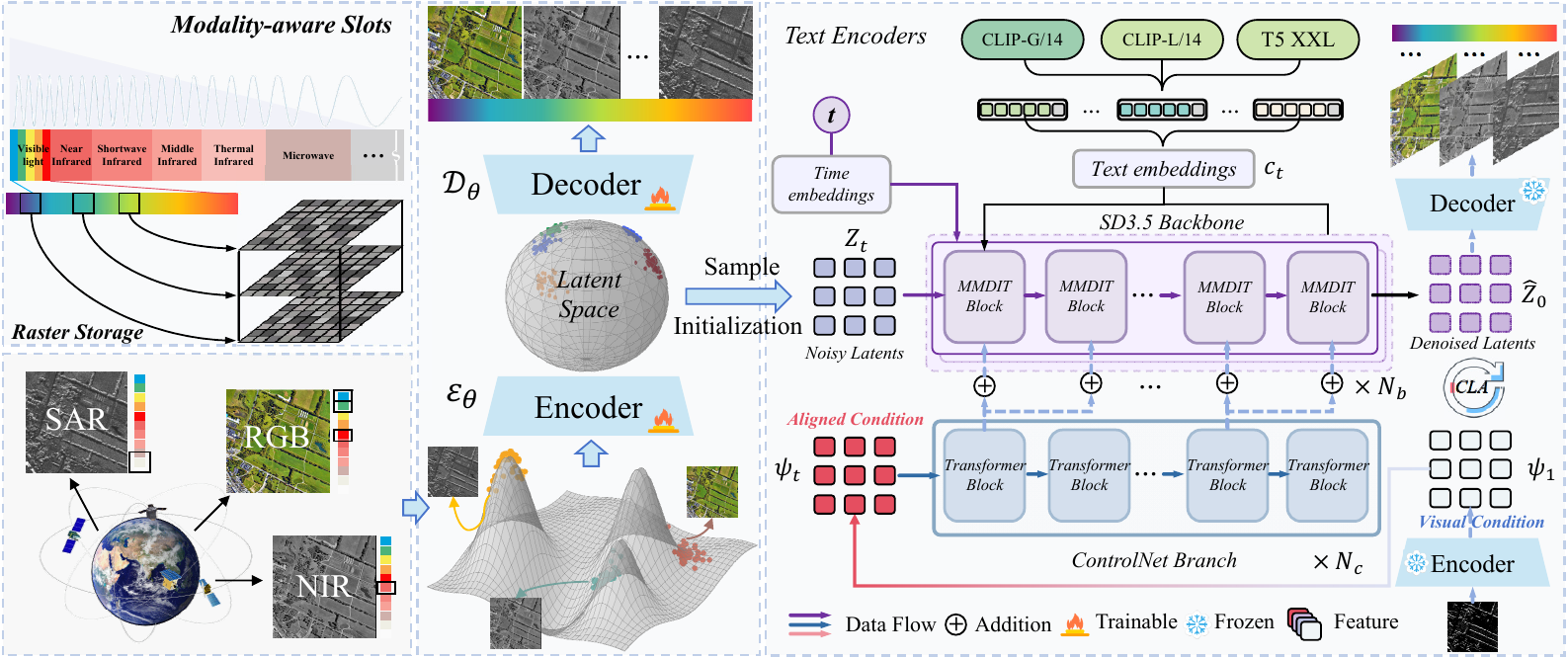}
	\vspace{-0.2em}
	\caption{Overview of OmniRSI. Multimodal observations are slot-organized and encoded into a unified latent space. SD3.5 backbone is fine-tuned for generation, with a ControlNet branch injecting hierarchical condition guidance. The derived latents are decoded for multimodal reconstruction.}
	\label{method:1-1}
\end{figure*}

OmniRSI aims to establish a unified model for multimodal remote sensing generation under heterogeneous and partially observed data (Fig.~\ref{method:1-1}). It is conceptualized into three vital functions, including multimodal image reconstruction, text-to-image generation and image-to-image translation. 

\begin{table*}[htp]
	\centering
	\caption{Details of RSI-MM datasets. \ding{51}: available;  --: unavailable; \ding{109}: supplemented by us.}\label{tab:RSI-MM}
	\vspace{-0.8em}
	\renewcommand\arraystretch{1.0}
	\resizebox{1.00\columnwidth}{!}{
		\begin{tabular}{c c c c c c | cccccc}
			\toprule
			\rowcolor{gray!5}  &  &  \multicolumn{4}{c|}{Modality} &  & &  &  &  & \\  
			\rowcolor{gray!5}\multirow{-2}{*}{{Datasets}}& \multirow{-2}{*}{Source\&Platform} &  RGB & MSI  & SAR & PAN &  \multirow{-2}{*}{Text} & \multirow{-2}{*}{Text Num.} &  \multirow{-2}{*}{Spat. Res.(m)}  & \multirow{-2}{*}{Band. Configs.}   & \multirow{-2}{*}{Image Num.}  & \multirow{-2}{*}{Image Size}   \\
			\midrule
			Git-10M & GoogleEarth\&Public datasets 
			& \ding{51}  &  --  &  -- &  --& \ding{51} & 10,452,649 & 0.5$\sim$128 & 3 & 10,452,649  & 256 $\times$ 256 \\
			EuroSAT & Sentinel-2
			& \ding{51} & \ding{51} &  -- &  -- & \ding{109}& 27,000 &  10/20/60  &  13 &  27,000 & 64 $\times$ 64 \\
			OSCD &  Sentinel-2   
			& \ding{51} & \ding{51} &  -- &  -- & \ding{109}& 48 & 10/20/60 &  13  &  48 & 241$\sim$1180  \\
			M3M-CR & PlanetScope\&Sentinel-1 
			& \ding{51} &  \ding{51}  & \ding{51} &  -- & \ding{109}& 63,000 & 3/10 &  5  & 63,000 & 300/90 \\
			OpenEarthMap-SAR & UmbraSpace\&NAIP\&IGN\&GSI
			& \ding{51} &  --  & \ding{51} &  -- & \ding{109}&  5,033 & 0.15-0.5 & 4 & 5,033 & 1024 $\times$ 1024	\\
			PIE-RGB-SAR &  Google Satellite\&GF-3	& \ding{51} &  -- & \ding{51} &  -- & \ding{109}& 4,865 & 0.5-3 & 4 &  4,865 & 256 $\times$ 256 \\
			WHU-OPT-SAR & GF-1\&GF-3& {\ding{51}} & {\ding{51}}  & {\ding{51}} &  -- & \ding{51} & 4,900 & 2/5 & 5  &  4,900   & 512$\times$ 512 \\ 
			SARLO  & Umbra\&Google Satellite & \ding{51} &  -- & \ding{51} &  -- & \ding{51}& 358,698 & 0.8 & 4 &  119,566  & 1024 $\times$ 1024\\
			RSCC  &	MAXAR	& \ding{51} &  -- &  -- &  -- & \ding{51}& 62,351  &  0.3$\sim$0.8  & 3 &  62,351    &  512 $\times$ 512 \\  
			RSICD &	GoogleEarth\&Baidu\&MapABC & \ding{51} &  -- &  -- &  -- & \ding{51} & 54,605 &  --  & 3 &  10,921    &  224 $\times$ 224 \\
			PSBench  & GF\&GE\&WV\&CBERS\&Landsat	& \ding{51} & \ding{51} &  -- & \ding{51} & \ding{109} & 449,936 & 0.65$\sim$30  & 5/8/9/11 &  449,936  & 256/512/1024  \\
			FSARCap &  GF3	&  -- &  -- & \ding{51} &  -- &\ding{51}&  72,400 &  1.0$\sim$5.0 &  1  &  14,480  & 1024 $\times$ 1024 \\ 
			GID & GF-2 & \ding{51} & \ding{51} & -- & -- & \ding{51} & 25,350 & 4.0 & 4 & 25,350 & 512$\times$ 512 \\
			Inria & -- & \ding{51} & -- & -- & -- & \ding{51}& 1,800 & 0.3 & 3 & 1,800 & 512 $\times$ 512 \\
			M4-SAR & Sentinel-1\&Sentinel-2 & \ding{51} & -- & \ding{51} & -- & \ding{109} & 112,174 & 10/60 & 4 & 112,174 & 512 $\times$ 512 \\
			ChatEarthNet & Sentinel-2 & \ding{51} & \ding{51} & -- & -- & \ding{51}& 169,220 & 10 & 9 & 164,656 & 256 $\times$ 256 \\
			\rowcolor{gray!20} RSI-MM(Total)   & Above datasets & \ding{51} & \ding{51} & \ding{51} & \ding{51}	& \ding{51} & 11,864,029	& 0.15$\sim$128	& 18	& 11,518,729 & 64$\sim$1180\\
			\bottomrule
	\end{tabular}}  
	\vspace{-0.8em}
\end{table*}

\textbf{Dataset Construction.} To fully support these tasks, we construct RSI-MM that contains 11.5 million multi-resolution images collected from diverse public remote sensing datasets, including EuroSAT~\cite{helber2018introducing}, GID~\cite{tong2020land}, Git-10M~\cite{liu2025text2earth}, OSCD~\cite{daudt2018urban}, M3M-CR~\cite{10354035}, OpenEarthMap-SAR~\cite{xia2025openearthmap}, PIE-RGB-SAR~\cite{zhang2024asanet}, Inria~\cite{maggiori2017can}, WHU-OPT-SAR~\cite{li2022mcanet}, SARLO~\cite{debuysere2025sarlo80}, RSCC~\cite{chen2025rscc}, RSICD~\cite{lu2017exploring}, M4-SAR~\cite{wang2025m4}, PSBench~\cite{wang2026universal}, FSARCap~\cite{zhang2026fsar}, ChatEarthNet~\cite{yuan2024chatearthnet}. As reported in Tab.~\ref{tab:RSI-MM}, these datasets cover heterogeneous sensors, spatial resolutions, and modality configurations, including RGB, MSI, SAR, and PAN. For datasets without text annotations, we select the representative bands to form visualizations and feed them to GPT-5.4-mini~\cite{singh2025openai} with a specific prompt to generate textual descriptions of scene information (See Appendix~\ref{sec:A}). 

\textbf{Multimodal Latent Representation Learning.} To accommodate heterogeneous and partially observed remote sensing data, we organize observations into $K$ predefined modality-aware slots according to their spectral characteristics and modality types (See Appendix~\ref{sec:B}). Let $\mathbf{X}=\{\mathbf{x}_k\}_{k=1}^{K}$ denote the slot-organized data and $\mathbf{M}=\{\mathbf{m}_k\}_{k=1}^{K}$ their binary mask, indicating the slot availability. The valid input is $\mathbf{X}_v=\mathbf{M}\odot\mathbf{X}$, with $\mathbf{M}$ broadcast over spatial dimensions and $\odot$ denoting element-wise multiplication. This organization provides a consistent data dimension across different sensors and modality combinations. Building on this scheme, we develop RS-VAE with an encoder $\mathcal{E}_{\theta}$ that maps $\mathbf{X}$ to a latent variable $\mathbf{Z}$ and defines the approximate posterior $q_{\theta}(\mathbf{Z}\!\mid\!\mathbf{X})$ and a decoder $\mathcal{D}_{\theta}$ reconstructing the slot-organized observations. It retains the standard VAE architecture~\cite{esser2024scaling} while adapting its input and output channels to the slot capacity $K$. To learn complementary multimodal representations and support reconstruction from incomplete observations, we jointly employ observed-slot and masked reconstruction constraints, with respect to $\mathcal{L}_{\mathrm{rec}}$ and $\mathcal{L}_{\mathrm{mask}}$:
\begin{equation}\label{eq:rsvae_reconstruction}
	\!\!\mathcal{L}_{\mathrm{rec}}\!=\!\mathbb{E}_{q_{\theta}(\mathbf{Z}\mid\mathbf{X}_v)}\!\left[\left\|\mathbf{M}\!\odot \!\left(\mathcal{D}_{\theta}(\mathbf{Z})\!-\!\mathbf{X}_v\right)\right\|_2^2\right]\!,\quad  \mathcal{L}_{\mathrm{mask}}\!= \!	\mathbb{E}_{q_{\theta}(\mathbf{Z}\mid \mathbf{X}_r)}\!\left[\left\|\mathbf{M}\!\odot\!\left(\mathcal{D}_{\theta}(\mathbf{Z})\!-\!\mathbf{X}_v\right)\right\|_2^2\right]\!,
\end{equation}
where $\mathbf{X}_r$ is obtained by randomly masking entire slots or spatial regions within individual slots. $\mathcal{L}_{\mathrm{rec}}$ captures and internalizes cross-modal dependencies by jointly reconstructing available observations, while $\mathcal{L}_{\mathrm{mask}}$ encourages recovery of withheld measurements from complementary modalities and spatial-spectral context. Furthermore, we promote locally smooth decoding by penalizing the sensitivity of $\mathcal{D}_{\theta}$ to symmetric latent perturbations:
\begin{equation}
	\label{eq:rsvae_smooth}
	\mathcal{L}_{\mathrm{smooth}}=\mathbb{E}_{\mathbf{Z},\boldsymbol{\eta}}\big[\|D_{\theta}(\mathbf{Z}+\boldsymbol{\eta})-D_{\theta}(\mathbf{Z}-\boldsymbol{\eta})\|_{1} / (\|\boldsymbol{\eta}\|_1+\delta) \big],
\end{equation}
where $\boldsymbol{\eta}\sim\mathcal{N}(0,\sigma_{\eta}^{2}\boldsymbol{I})$ is a small Gaussian perturbation and $\delta$ is a small positive value that ensures numerical stability. This constraint discourages abrupt changes in reconstruction under local latent variations. Together with the posterior regularization
$\mathcal{L}_{\mathrm{KL}}=\mathbb{E}_{q_{\theta}(\mathbf{Z}\mid\mathbf{X})}[D_{\mathrm{KL}}(q_{\theta}(\mathbf{Z}\!\mid\!\mathbf{X})
\| \mathcal{N}(0,\boldsymbol{I}))]$ that constrains representation toward an isotropic Gaussian prior, the overall objective is defined as:
\begin{equation}
	\label{eq:rsvae_total}
	\mathcal{L}_{\mathrm{vae}} =\mathcal{L}_{\mathrm{rec}}+\alpha\mathcal{L}_{\mathrm{mask}}+\gamma\mathcal{L}_{\mathrm{smooth}}+	\beta\mathcal{L}_{\mathrm{KL}},
\end{equation}
where $\alpha$, $\gamma$, and $\beta$ balance each constraint. These four complementary objectives establish a shared representation space for unified multimodal generation and translation.
 
\textbf{Multimodal Generation and Translation.} Grounded in the unified latent space, we adopt SD3.5 Medium with $N_b$ MMDiT blocks as the backbone for text-to-image generation (See Appendix~\ref{sec:C}). Specifically, text prompts are jointly encoded by CLIP-G/14, CLIP-L/14~\cite{beaumont2025large}, and T5-XXL~\cite{raffel2020exploring}, collectively yielding the textual condition $c$. Meanwhile, the pretrained encoder $\mathcal{E}_\theta$ is frozen and maps the available data to a target representation $\mathbf{Z}_0 = \mathcal{E}_\theta(\mathbf{X})$. Following the Flow matching~\cite{lipman2023flow,liu2022rectified} paradigm, given $\mathbf{Z}_0$ and independent Gaussian noise $\mathbf{Z}_1\!\sim\!\mathcal{N}(0,\boldsymbol{I})$, an intermediate latent state $\mathbf{Z}_t$ is constructed as $\mathbf{Z}_t=(1-\sigma_t)\mathbf{Z}_0+\sigma_t \mathbf{Z}_1$, where $\sigma_t\in[0,1]$ specifies the interpolation schedule. The model learns a conditional velocity field $v_{\theta}(\mathbf{Z}_t,t,c)$ by regressing the sample-wise target velocity:
\begin{equation}\label{eq:rs_flow_matching}
	\mathcal{L}_{\mathrm{FM}}=\mathbb{E}_{\mathbf{Z}_0,\mathbf{Z}_1,t,c}\left[	\left\|	v_{\theta}(\mathbf{Z}_t,t,c)-\dot{\sigma}_t
	(\mathbf{Z}_1-\mathbf{Z}_0)\right\|_2^2\right],
\end{equation}
where $\dot{\sigma}_t=d\sigma_t/dt$. At inference, one can utilize the classifier-free guidance to steer generative trajectories by combining unconditional and conditional velocity predictions:
\begin{equation}\label{sec:method-fm2}
	v_{\mathrm{cfg}}(\mathbf{Z}_t,t,c)=v_{\theta}(\mathbf{Z}_t,t,\varnothing)+w\left[v_{\theta}(\mathbf{Z}_t,t,c)-v_{\theta}(\mathbf{Z}_t,t,\varnothing)\right],
\end{equation}
where $\varnothing$ denotes the null text condition and $w$ controls the guidance strength. Starting from stochastic noise $\mathbf{Z}_1\sim\mathcal{N}(0,\boldsymbol{I})$, an ordinary differential equation solver integrates
$d\mathbf{Z}_t/dt=v_{\mathrm{cfg}}(\mathbf{Z}_t,t,c)$
from $t=1$ to $t=0$ to obtain the generated latent $\hat{\mathbf{Z}}_0$. Then the frozen decoder $\mathcal{D}_{\theta}$ maps back to the pixel space as $\hat{\mathbf{X}}=\mathcal{D}_{\theta}(\hat{\mathbf{Z}}_0)$, producing full-slot outputs across heterogeneous modalities.
 
To support conditional translation from observed data, we augment the backbone with a trainable ControlNet branch containing $N_c$ Transformer blocks. The observed data $\mathbf{X}$ are first projected into the latent space to obtain the condition as $\boldsymbol{\psi}=\mathcal{E}_{\theta}(\mathbf{X})$, which is then processed by the ControlNet branch to produce hierarchical control features $\{\mathbf{h}^{c}_{j}\}_{j=1}^{N_c}$. These features are residually added to the hidden states $\mathbf{H}_i$ of the SD3.5 backbone after each MMDiT block, which can be expressed as:
\begin{equation}\label{eq:control_injection}
	\mathbf{H}_i\leftarrow\mathbf{H}_i+\mathbf{h}^{c}_{g(i)},\qquad	g(i)=1+\left\lfloor	N_c\cdot (i-1) / N_b \right\rfloor,\quad i=1,\ldots,N_b,
\end{equation}
where $\lfloor\cdot \rfloor$ denotes the floor operation. For $N_c\leq N_b$, this assignment distributes control features across consecutive groups of backbone blocks, providing visual guidance at multiple depths. Accordingly, each control feature modulates a group of consecutive MMDiT blocks, allowing the partially observed modalities to provide hierarchical visual guidance throughout generation. Moreover, both of RS-VAE and SD3.5 backbone are frozen, while the additional branch is optimized with the similar objective in Eq.~(\ref{eq:rs_flow_matching}) using the conditional velocity $v_{\theta}(\mathbf{Z}_t,t,c,\boldsymbol{\psi})$. This formulation supports translation from different subsets of observed modalities within a shared generative model.

\begin{figure*}[t]
	\centering 
	\begin{minipage}[t]{0.59\textwidth}
		\begin{algorithm}[H]\label{alg:cla}
			\caption{Cyclic Latent Alignment}
			\KwIn{$\mathbf{X}$,  $\mathbf{M}$, $c$, $\mathcal{E}_{\theta}$, $\mathcal{D}_{\theta}$,$v_{\theta}$, $\{\lambda_t\}$, $\{\sigma_t\}$.}
			
			\phantom{******}  $\mathbf{Z}_1 \sim \mathcal{N}(0,\mathbf{I})$ , $\boldsymbol{\psi}_{1}= \mathcal{E}_{\theta}(\mathbf{M}\odot \mathbf{X})$
			
			\For{$t=1,1-\Delta t,\ldots,\Delta t$}{
				
				$v_t \leftarrow v_{\theta}(\mathbf{Z}_t,t,c,\boldsymbol{\psi}_t)$
				 
				$\widehat{\mathbf{Z}}_0^{t} \leftarrow \mathbf{Z}_t-(\sigma_t-\sigma_0)v_t$\
				
				$\boldsymbol{\psi}^{\ast} = \mathbf{M} \odot \mathbf{X} + (1-\mathbf{M}) \odot \mathcal{D}_\theta(\widehat{\mathbf{Z}}^{t}_{0})$
				
				$\boldsymbol{\psi}_{t-\Delta t} = \lambda_t \boldsymbol{\psi}_{1} + (1-\lambda_t) \mathcal{E}_\theta(\boldsymbol{\psi}^{\ast})$
						 
				$\mathbf{Z}_{t-\Delta t} \leftarrow \mathbf{Z}_{t} - (\sigma_{t} -\sigma_{t-\Delta t})v_{\theta}(\mathbf{Z}_{t},t,c,\boldsymbol{\psi}_{t-\Delta t})$
				
			}
 			
 			$\mathbf{X} \leftarrow \mathcal{D}_{\theta}(\mathbf{Z}_0) $
			
			\Return{$\mathbf{X}$}
		\end{algorithm}
	\end{minipage}
	\hfill
	\begin{minipage}[t]{0.40\textwidth}
		\vspace{0pt}
		\centering
		\includegraphics[width=\linewidth]{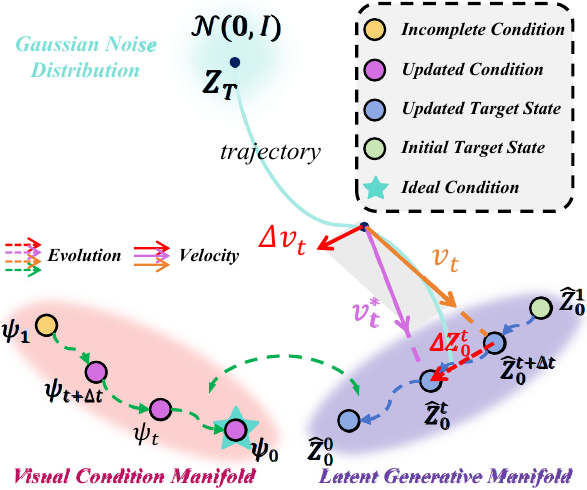}
		\vspace{-2.0em}
		\caption{Cyclic Latent Alignment.}
		\label{fig:cla_pipeline}
	\end{minipage}
\end{figure*}

\begin{figure*}[h]
	\centering 
	\begin{minipage}{\textwidth}
	\centering
	\includegraphics[width=0.99\linewidth]{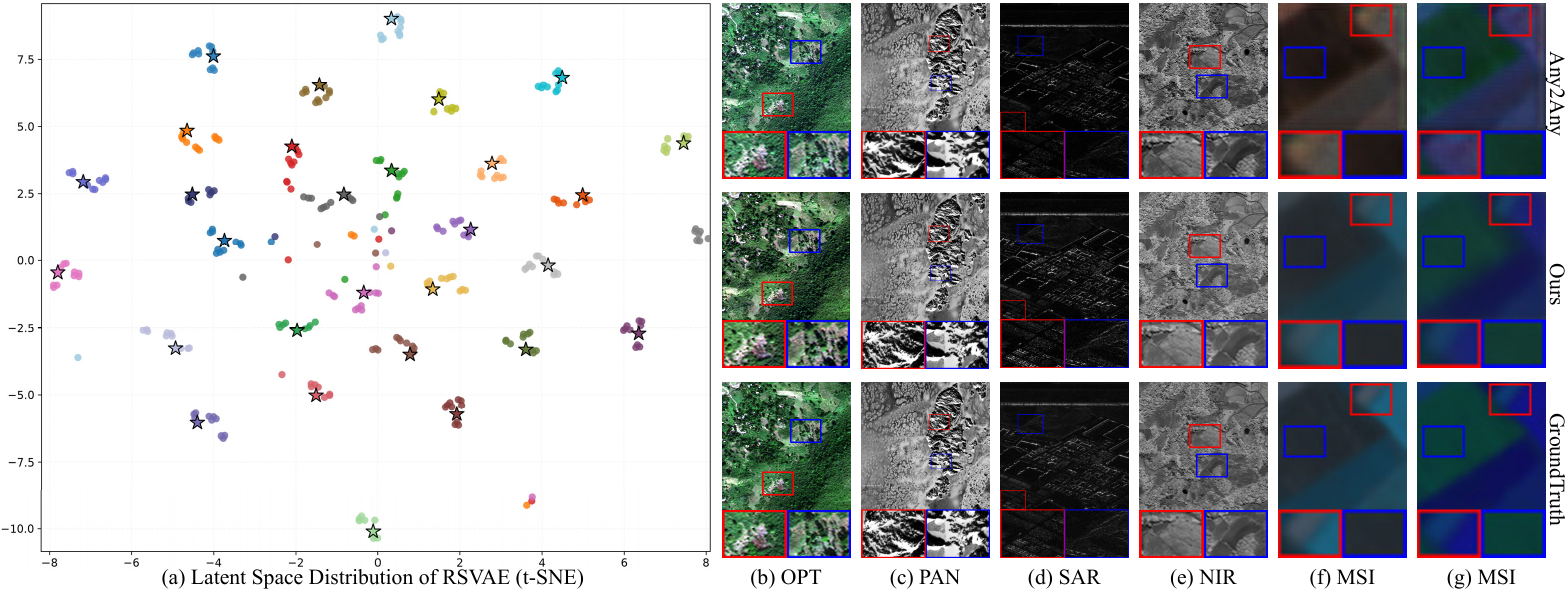}
	\vspace{-0.1\floatsep} 	
	\caption{Representation and reconstruction of RS-VAE. (a) Latent distribution, where stars denote different samples and circles are their corresponding subset bands. Most representations cluster tightly around their associated sample, indicating that RS-VAE preserves scene-level consistency well for heterogeneous observations. (b)-(g) Reconstruction comparisons across OPT, PAN, SAR, near-infrared (NIR), and MSI modalities, including two Sentinel-2 composites (B5/B6/B7 and B8A/B11/B12). The top/middle/bottom rows are results of Any2any, RS-VAE, groundtruth (GT).}
	\label{fig:vae}
	\end{minipage}
	
	\begin{minipage}{\textwidth}
	 \centering
		\includegraphics[width=0.99\linewidth]{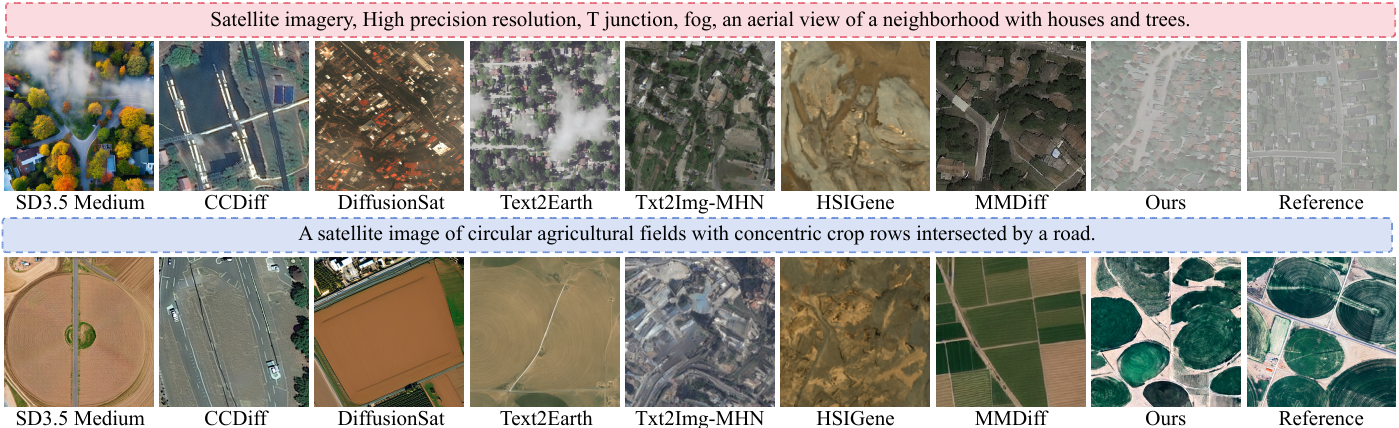}
		\vspace{-0.4em}
		\caption{Qualitative results of text-to-image generation on RGB composites.}
		\label{fig:t2i}
	\end{minipage} 
\end{figure*}

\textbf{Cyclic Latent Alignment}\label{sec:vla}
Textual descriptions primarily convey global semantics but remain inherently sparse for wide-field-of-view remote sensing images, where fine-grained spatial and spectral details resist exhaustive specification. Partially observed modalities offer more explicit visual guidance, but their incompleteness may bias conditional velocity predictions. Given partially observed data $\mathbf{X}$ with availability mask $\mathbf{M}$, the coarse visual condition is obtained by $\boldsymbol{\psi}_{1}=\mathcal{E}_\theta(\mathbf{M}\odot\mathbf{X})$. The Euler solver is then used to approximate the solution, with the forward and reverse steps given by:
\begin{equation}\label{sec:method-cla1}
	\textbf{Fwd:}\, \mathbf{Z}_{t+\Delta t}\!  = \! \mathbf{Z}_t \! +\! v(\mathbf{Z}_t,t,c,\boldsymbol{\psi}_0)(\sigma_{t+\Delta t}\!-\! \sigma_t), \, \textbf{Inv:}\,\mathbf{Z}_{t-\Delta t}\!  =\!  \mathbf{Z}_{t} \!-\! v_\theta(\mathbf{Z}_{t},t,c,\boldsymbol{\psi}_{1})(\sigma_{t} \!-\! \sigma_{t-\Delta t}),
\end{equation}
where timestep $t$ increases from $0$ to $1$ in the forward process and decreases from $1$ to $0$ in the reverse process, advancing by $\Delta t$ at each step. Evidently, the incomplete latent variable $\boldsymbol{\psi}_{1}$ is available in the reverse process, while the desired latent $\boldsymbol{\psi}_0$ remains unknown at each step. Consequently, the velocity field $v_\theta(\mathbf{Z}_t,t,c,\boldsymbol{\psi}_{0})$ cannot be directly evaluated, making the exact generation intractable. In this situation, we introduce a training-free CLA sampling strategy, which progressively enriches incomplete visual conditions using generated multimodal information and iteratively corrects the conditional flow trajectory. Specifically, at the $t$-th alignment cycle, the model predicts the velocity field $v_{t}=v_{\theta}(\mathbf{Z}_{t},t,c,\boldsymbol{\psi}_t)$, and numerically integrates from current state to yield the one-step generated endpoint state $\widehat{\mathbf{Z}}^{t}_{0}$. Instead of recycling the estimate as the visual condition, we decode it and fill only missing entries, preserving observed measurements. Then the enriched observation is re-encoded and interpolated with the initial condition $\boldsymbol{\psi}_{1}$:
\begin{equation}
	\boldsymbol{\psi}_{t-\Delta t} = \lambda_t \boldsymbol{\psi}_{1} + (1-\lambda_t) \mathcal{E}_\theta(\boldsymbol{\psi}^{\ast}) \leftarrow \boldsymbol{\psi}^{\ast} = \mathbf{M} \odot \mathbf{X} + (1-\mathbf{M}) \odot \mathcal{D}_\theta(\widehat{\mathbf{Z}}^{t}_{0}),
\end{equation}
where $\lambda_t$ is the annealed coefficient that decreases from $1$ to $0$, and $\boldsymbol{\psi}_{t-\Delta t}$ is the updated visual condition. This schedule exploits the generated estimate to compensate for missing modalities, while progressively re-anchoring the condition to the generated evidence as sampling approaches the data manifold. Such interpolation is motivated by the observed proximity of latent representations under different band configurations (Fig.~\ref{fig:vae}) and the observation that richer visual conditions generally improve translation fidelity (Appendix~\ref{sec:D-2}). Then the reverse trajectory is advanced by:
\begin{equation}
	\mathbf{Z}_{t-\Delta t} = \hat{\mathbf{Z}}_{t-\Delta t} - (\sigma_{t} -\sigma_{t-\Delta t})(v_{t}^{\ast} - v_{t}) \xleftarrow{v_{t}^{\ast}=v_{\theta}(\mathbf{Z}_{t},t,c,\boldsymbol{\psi}_{t-\Delta t})} \hat{\mathbf{Z}}_{t-\Delta t} = \mathbf{Z}_t - (\sigma_{t} -\sigma_{t-\Delta t}) v_{t},
\end{equation}
By repeated alignment, CLA continuously refines the visual condition and steers the transport direction, yielding more reliable velocity estimates without additional training.  Each step requires two velocity evaluations, one for endpoint estimation and one for trajectory advancement under the updated condition. The total number of neural function evaluations (NFE) is $2/\Delta t$, scaling linearly with the sampling steps. The detailed algorithm is summarized in Alg.~\ref{alg:cla} and presented in Fig.~\ref{fig:cla_pipeline}.
 
\section{Experiments}  

\textbf{Experimental Setup.} RSI-MM consists of 16 public datasets, from each of which we select a subset of samples for task evaluation (See Appendix~\ref{sec:C-1}). For image reconstruction, we compare RS-VAE against Any2any~\cite{chen2026any2any} with five modality-specific VAEs using PSNR and SSIM~\cite{wang2004image}; For text-to-image generation, due to the lack of unified methods and metrics across remote sensing modalities, we compare OmniRSI with representative methods including Txt2Img-MHN~\cite{xu2023txt2img}, CC-Diff~\cite{zhang2024cc}, DiffusionSat~\cite{khanna2024diffusionsat}, SD3.5 Medium~\cite{esser2024scaling}, MMDiff~\cite{tang2026mmdiff}, HSIGene~\cite{pang2026hsigene}, Text2Earth~\cite{liu2025text2earth}, all using their official pretrained weights. Generation quality is evaluated by FID~\cite{heusel2017gans}, NIQE~\cite{mittal2012making}, CLIPIQA~\cite{wang2023exploring}, and CLIPScore~\cite{hessel2021clipscore} on RGB bands; For multimodal translation, we report PSNR and SSIM and compare OmniRSI with DDPM~\cite{songscore}, ReFlow~\cite{liuflow}, DDBM~\cite{zhou2024denoising}, RDDM~\cite{liu2024residual}, and RDBM~\cite{wang2026residual}. These baselines are adapted by modifying their input/output dimensions to match slot number and retrained separately for each task.
 
\textbf{Implementation Details.} OmniRSI is implemented in PyTorch and trained in three stages. Firstly, RS-VAE is optimized using $\mathcal{L}_{vae}$ with empirically selected weights $\alpha\!=\!0.1,\beta\!=\!0.00025,\gamma\!=0.01\!$. Secondly, SD 3.5 Medium is initialized from pretrained weights and its MMDiT blocks ($N_b\! = \!24$) are fully fine-tuned, while RS-VAE and text encoders (\emph{i.e., CLIP-G/14, CLIP-L/14, and T5-XXL}) remain frozen. Thirdly, the ControlNet branch with $N_c \!= \!12$ Transformer blocks is initialized from SD3.5 backbone weights and trained with all other components frozen. The coefficient schedules are $\sigma_t \!=\! \lambda_t \!=\! t$. All training uses AdamW with a learning rate of $3\!\times\!10^{-5}$ under cosine decay and proper batch size. Inference uses 40 steps and default seeds, with $\omega=5.0$ for generation and $1.0$ for translation. The total training takes 60 days on 8 NVIDIA A100 80GB GPUs (See Appendix~\ref{sec:C-2}).
 
\begin{table}[t]
	\caption{Quantitative results of our RS-VAE and modality-specific VAEs in Any2any. FLOPS are measured at 256$\times$256 resolution. \textbf{Bold} ranks the best and \underline{underline} ranks the second best.}\label{tab:vae}
	\vspace{-0.1\floatsep} 	
	\centering
	\renewcommand\arraystretch{1.1}
	\resizebox{1.0\columnwidth}{!}{
		\begin{tabular}{c|cc|cc|cc|cc|cc|cc|cc}
			\toprule
			\multirow{2}{*}{\textbf{Method}}  & \multicolumn{2}{c|}{PAN} & \multicolumn{2}{c|}{OPT} & \multicolumn{2}{c|}{SAR} & \multicolumn{2}{c|}{NIR} & \multicolumn{2}{c|}{MSI} &  \multicolumn{2}{c|}{Average} &  \multicolumn{2}{c}{Complexity} \\ 
			& PSNR$\uparrow$ & SSIM$\uparrow$ & PSNR$\uparrow$ & SSIM$\uparrow$ & PSNR$\uparrow$ & SSIM$\uparrow$ & PSNR$\uparrow$ & SSIM$\uparrow$ & PSNR$\uparrow$ & SSIM$\uparrow$ & PSNR$\uparrow$ & SSIM$\uparrow$ & Params(M) & FLOPs(G)\\
			\midrule 
			Any2any & \underline{37.68}  & \underline{0.838}  & \underline{39.06}  & \underline{0.951}  & \underline{26.85}  & \underline{0.846}  & \underline{29.99}  & \underline{0.951}  & \underline{21.78}  & \underline{0.796} & \underline{32.69} & \underline{0.890} & 13.322$\sim$83.610 &  329.120$\sim$473.663 \\
			Ours    & \textbf{42.79}  & \textbf{0.972}  & \textbf{39.71}  & \textbf{0.958}   & \textbf{27.15}  & \textbf{0.878} & \textbf{32.83}  & \textbf{0.953} & \textbf{26.91}  & \textbf{0.869} & \textbf{34.95} &  \textbf{0.935} &  83.815 & 447.742 \\	 
			\bottomrule
	\end{tabular}}
	\vspace*{-1.3em}
\end{table} 

\begin{table*}[t]
	\centering 
	\begin{minipage}{\textwidth}
		\centering
		\begin{minipage}[t]{0.49\textwidth}
			\caption{Results of text-to-image task.}\label{tab:t2i}
			\vspace{-1.0em}
			\centering
			\renewcommand\arraystretch{1.0}
			\resizebox{\linewidth}{!}{%
			\begin{tabular}{c|cccc}
				\toprule
				\multirow{1}{*}{\textbf{Method}}   & FID Score $\downarrow$   & NIQE$\downarrow$ & CLIPIQA$\uparrow$  & CLIPScore$\uparrow$  \\  
				\midrule   
				CC-Diff       & 132.137  & 6.168  & 0.569 & 0.725   \\ 
				Txt2Img-MHN   & 150.063  & \underline{5.398}  & 0.377 & 0.672 \\
				HSIGene       & 105.162  & 7.463  & 0.217 & 0.662    \\
				Text2Earth    & \underline{52.842} & 7.362  & 0.471 & 0.750   \\
				DiffusionSat  & 81.799   & 6.912  & 0.541   & 0.755 \\
				MMDiff        & 80.922   & 5.824  & 0.377   & \underline{0.780}   \\
				SD3.5 Medium  & 93.089   & 6.397  & \underline{0.597} & 0.758   \\
				Ours          & \textbf{28.693}   & \textbf{4.399}  & \textbf{0.695} & \textbf{0.830}  \\
				\bottomrule
			\end{tabular}}
		\end{minipage}
		\hfill
		\begin{minipage}[t]{0.49\textwidth}
			\caption{Ablation on varied $\omega$ value.}\label{tab:guidance}
			\vspace{-1.0em}
			\centering
			\renewcommand\arraystretch{1.18}
			\resizebox{\linewidth}{!}{%
				\begin{tabular}{c|cccc}
 					\toprule
					\multirow{1}{*}{\textbf{Guidance}}  & FID Score $\downarrow$  & NIQE$\downarrow$ & CLIPIQA$\uparrow$  & CLIPScore$\uparrow$  \\  
					\midrule  
					$\omega = 1.0$  & 42.077  & 4.790  & 0.602 & 0.786   \\ 
					$\omega = 2.0$  & 35.084  & \underline{4.767}  & 0.639 & 0.800    \\
					$\omega = 3.0$  & 30.908  & 4.793  & 0.633 & \underline{0.805}   \\
					$\omega = 4.0$  & \underline{29.160}  & 4.864  & \underline{0.668} & 0.803    \\
					$\omega = 5.0$  & \textbf{28.693}  & \textbf{4.399}  & \textbf{0.695} & \textbf{0.830}   \\
					$\omega = 6.0$  & 29.183  & 4.924  & 0.661 & 0.802   \\ 
					\bottomrule
			\end{tabular}}
		\end{minipage}
	\end{minipage}
	
	\vspace{0.5\floatsep}
	
	\begin{minipage}{\textwidth}
	 	\caption{Quantitative comparisons of translation tasks.}\label{tab:i2i}
	 	\vspace{-1.0em} 
		\centering
		\renewcommand\arraystretch{1.}
		\resizebox{\linewidth}{!}{%
			\begin{tabular}{c|cc|cc|cc|cc|cc|cc}
 				\toprule
				\multirow{2}{*}{\textbf{Method}}  & \multicolumn{2}{c|}{PAN $\rightarrowtail$ MSI} & \multicolumn{2}{c|}{MSI $\rightarrowtail$ PAN} & \multicolumn{2}{c|}{SAR $\rightarrowtail$ MSI} & \multicolumn{2}{c|}{MSI $\rightarrowtail$ SAR} &  \multicolumn{2}{c|}{Average} &  \multicolumn{2}{c}{Complexity} \\ 
				& PSNR$\uparrow$ & SSIM$\uparrow$ & PSNR$\uparrow$ & SSIM$\uparrow$ & PSNR$\uparrow$ & SSIM$\uparrow$ & PSNR$\uparrow$ & SSIM$\uparrow$ & PSNR$\uparrow$ & SSIM$\uparrow$ & Params(M) & FLOPs(G)\\
				\midrule  
				DDPM  & 10.88  & 0.271  & 12.49  & 0.335   & \underline{12.23}  & 0.126  & 12.81  & 0.057  & 12.01 & 0.232   & 9.24  & 34.245\\
				ReFlow& 18.72  & 0.564  & 29.71  & 0.813   & 10.16  & 0.239  & 15.46  & 0.196  & 20.62 & 0.533   & 34.25 & 9.238\\
				DDBM  & 16.35  & 0.020  & 20.99  & 0.123   & 8.45   & 0.015   & 9.92  & 0.032  & 15.55 & 0.058   & 30.09 & 64.305\\
				RDDM  & 19.73  & 0.360  & 29.92  & 0.612   & 12.13  & 0.253  & 16.99  & 0.118  & 21.59 & 0.389   & 60.41 & 7.495     \\   
				RDBM  & \underline{21.02}  & \underline{0.656}  & \underline{31.11}  & \underline{0.881} & 12.16  & \underline{0.367}  & \underline{18.01}  & \underline{0.245}  & \underline{22.59} & \underline{0.616}   & 39.17 & 7.83 \\  
				\hline
				Ours(\ding{53}text) & \textbf{22.07} & \textbf{0.662} & \textbf{35.17} & \textbf{0.914} & \textbf{17.36} & \textbf{0.373} & \textbf{20.32} & \textbf{0.291} & \textbf{25.63} & \textbf{0.639} & 2789.06 & 1968.955 \\	  
				Ours(\ding{51}text) & 25.16 & 0.786 & 36.10 & 0.925 & 18.39 & 0.382 & 20.50 & 0.302 & 27.10 & 0.683 & 2789.06 & 1968.955 \\
				\bottomrule
		\end{tabular}}
	\end{minipage}
	\vspace*{-0.8em}
\end{table*}

\begin{figure*}[t]
	\centering
	\includegraphics[width=0.99\linewidth]{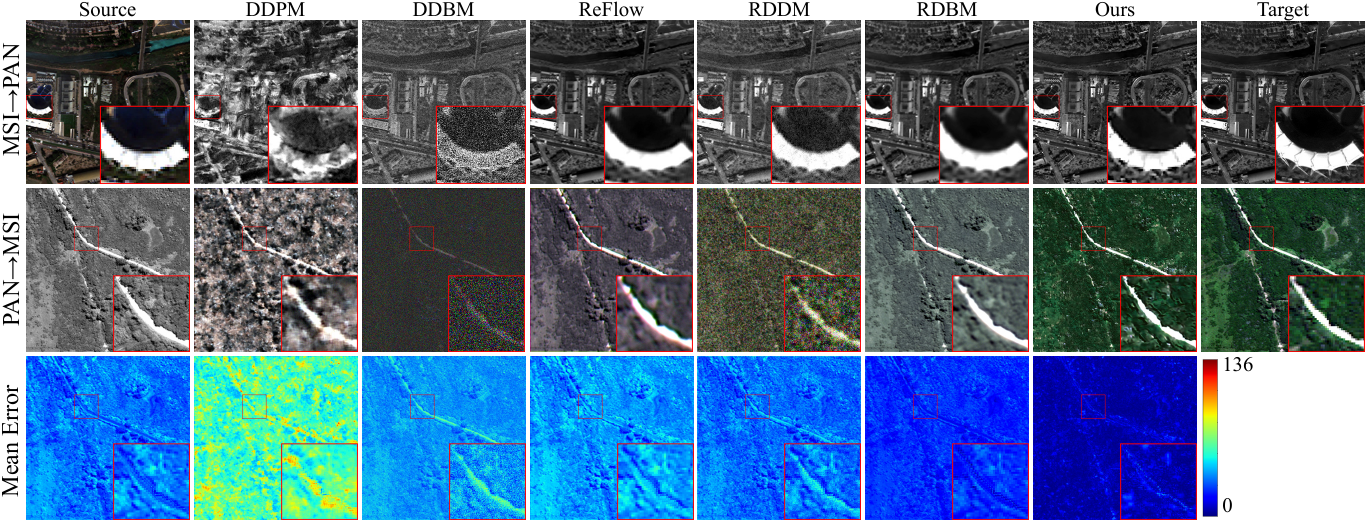}
	\vspace{-0.2em}
	\caption{Qualitative results of image-to-image generation. Displayed in RGB composite.}
	\label{fig:i2i}
\end{figure*}

\subsection{Main Results}
\textbf{Image Reconstruction.} RS-VAE aims to map heterogeneous observations into a shared latent space. As presented in Fig.~\ref{fig:vae}, full-observation representations overlap with multiple single-band regions, reflecting the integration of complementary modality information. Following task settings of Any2any, RS-VAE consistently achieves the best reconstruction performance across specific modality combinations in Tab.~\ref{tab:vae}. Further band-wise evaluation in Appendix~\ref{sec:D-0} demonstrates robust self-reconstruction from sparse observations and progressively improved full-band reconstruction with increasing band availability, confirming its effective exploitation of modality dependencies.

\textbf{Text-to-image generation.} For fairness, we use identical text prompts and compare RGB composites extracted from their outputs, since existing methods are typically adapted to modality-specific datasets and output spaces. As illustrated in Fig.~\ref{fig:t2i}, OmniRSI produces images with improved semantic alignment, coherent spatial layouts, and faithful scene structure. Quantitative results in Tab.~\ref{tab:t2i} further confirm that OmniRSI achieves the best performance across all four metrics. Moreover, the performance of varied guidance strength in Tab.~\ref{tab:guidance} shows that $\omega=5.0$ provides the best overall trade-off, while overly weak or strong guidance degrades generation quality. Beyond RGB synthesis, we provide the band-wise visualizations for other modalities in Fig.~\ref{fig:t2ifull} and Fig.~\ref{fig:t2ifull2} (See Appendix~\ref{sec:D-1}), where the generated data are compared with available ground-truth bands and extended bands absent from the reference data. These results indicate that the unified latent space supports coherent multimodal synthesis despite incomplete pairing and partial observations during training, extending OmniRSI beyond specific synthesis toward unified multimodal generation.
 
\textbf{Multimodal Image Translation.} We assess translation under two representative tasks, PAN$\leftrightarrow$MSI and SAR$\leftrightarrow$MSI, where MSI encompasses both RGB and satellite-specific bands, as illustrated in Fig.~\ref{fig:i2i} and Fig.~\ref{fig:i2i2}. OmniRSI produces the most faithful cross-modal translations, preserving spatial structures while recovering modality-specific characteristics across PAN, MSI, and SAR observations. Quantitative results in Tab.~\ref{tab:i2i} further confirm this advantage. Despite using a unified model for all modality mappings, OmniRSI achieves the highest average PSNR and SSIM across all translation tasks and substantially outperforms other methods separately optimized for individual task, demonstrating the effectiveness of unified latent representation and powerful generation capacity. The effects of textual conditioning and guidance strength $\omega$ are reported in Tab.~\ref{tab:ai2i}. Without text prompts, the conditional and unconditional branches in Eq.~\ref{sec:method-fm2} share identical text embeddings and visual conditions, causing the guidance residual to vanish and rendering $\omega$ ineffective. Textual descriptions provide complementary semantic guidance to visual observations, substantially improving performance, while task-specific guidance strengths further enhance image fidelity. To evaluate the spectral fidelity, we evaluate the generated MSI images via four representative normalized-difference indices: NDVI~\cite{tucker1979red}, NDWI~\cite{mcfeeters1996use}, NDRE~\cite{barnes2000coincident}, and NDBI~\cite{zha2003use}, which characterize vegetation, water, red-edge, and built-up information, respectively. These indices are computed using the corresponding spectral bands available under each sensor configuration (See Appendix~\ref{sec:D-3}), as shown in Tab.~\ref{tab:ndxi}, OmniRSI achieves the best metrics across all indices, indicating more accurate preservation of inter-band spectral relationships. Fig.~\ref{fig:ndvi} further shows that the derived index maps more closely resemble the references, whereas competing methods exhibit noticeable spectral distortions. Since PAN-SAR pairs are unavailable in RSI-MM, we perform PAN$\leftrightarrow$SAR translation in a zero-shot manner, as shown in Fig.~\ref{fig:pan2sar} (See Appendix~\ref{sec:D-2}). These visualizations demonstrate that OmniRSI can transfer across unseen modality pairs without task-specific adaptation, further highlighting its competitive generalization capacities.
 
\begin{table*}[t]
	\begin{minipage}[t]{0.73\textwidth} 
		\vspace{0pt}
		\centering
		\caption{Ablation study on the effect of $\omega$ for different translation tasks.}\label{tab:ai2i} 
		\vspace{-1.0em}
		\renewcommand\arraystretch{1.1}
		\resizebox{1.0\linewidth}{!}{%
			\begin{tabular}{c|cc|cc|cc|cc|cc}
				\toprule
				\multirow{2}{*}{\textbf{Guidance}} & \multicolumn{2}{c|}{PAN $\rightarrowtail$ MSI} & \multicolumn{2}{c|}{MSI $\rightarrowtail$ PAN} & \multicolumn{2}{c|}{SAR $\rightarrowtail$ MSI} & \multicolumn{2}{c|}{MSI $\rightarrowtail$ SAR} &  \multicolumn{2}{c}{Average}\\ 
				& PSNR$\uparrow$ & SSIM$\uparrow$ & PSNR$\uparrow$ & SSIM$\uparrow$ & PSNR$\uparrow$ & SSIM$\uparrow$ & PSNR$\uparrow$ & SSIM$\uparrow$ & PSNR$\uparrow$ & SSIM$\uparrow$\\
				\hline
				\multicolumn{11}{l}{\textbf{\ding{53} Text prompts, \ding{51} Euler, \ding{51} CLA}}\\
				\midrule  
				$\omega = 1.0$ & 22.07 & 0.662 & 35.17 & 0.914 & 17.36 & 0.373 & 20.32 & 0.291 & 25.63 & 0.639 \\
				$\omega = 2.0$ & 22.07 & 0.662 & 35.17 & 0.914 & 17.36 & 0.373 & 20.32 & 0.291 & 25.63 & 0.639 \\
				$\omega = 3.0$ & 22.07 & 0.662 & 35.17 & 0.914 & 17.36 & 0.373 & 20.32 & 0.291 & 25.63 & 0.639 \\
				$\omega = 4.0$ & 22.07 & 0.662 & 35.17 & 0.914 & 17.36 & 0.373 & 20.32 & 0.291 & 25.63 & 0.639 \\
				\hline
				\multicolumn{11}{l}{\textbf{\ding{51} Text prompts, \ding{51} Euler, \ding{51} CLA}}\\
				\hline
				\rowcolor{blue!5} $\omega = 1.0$ & \textbf{25.16} & 0.786 & \textbf{36.10} & \textbf{0.925} & \textbf{18.39} & \underline{0.382} & \textbf{20.50} & \textbf{0.302} & \textbf{27.10} & \underline{0.683} \\
				$\omega = 2.0$ & \underline{25.08} & \textbf{0.811} & \underline{35.87} & \underline{0.922} & \underline{18.25} & \textbf{0.390} & \underline{20.32} & \underline{0.291} & \underline{26.94} & \textbf{0.689} \\
				$\omega = 3.0$ & 24.27 & \underline{0.801} & 35.56 & 0.918 & 16.46 & 0.351 & 20.30 & 0.291 & 26.28 & 0.678 \\
				$\omega = 4.0$ & 23.35 & 0.776 & 35.06 & 0.912 & 16.19 & 0.341 & 20.22 & 0.291 & 25.76 & 0.666 \\  
				\bottomrule
		\end{tabular}}
	\end{minipage}
	\hfill
	\begin{minipage}[t]{0.26\textwidth}
		\vspace{0pt}
		\centering
		\includegraphics[width=\linewidth]{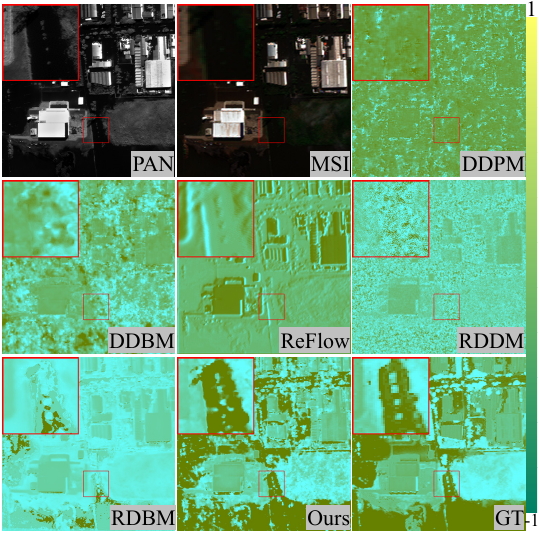}
		\vspace{-2.0em}
		\captionof{figure}{NDVI Results.}
		\label{fig:ndvi}
	\end{minipage}
	
	\vspace{0.5\floatsep}
	
	\begin{minipage}{\textwidth}
		\centering
		\renewcommand\arraystretch{0.95}
		\caption{Quantitative results of NDXI application across diverse translated MSI modalities.}
		\vspace{-1.0em}
		\label{tab:ndxi}
		\resizebox{1.0\columnwidth}{!}{
			\begin{tabular}{c|c@{\,~}@{\,~}c@{\,~}@{\,~}|c@{\,~}@{\,~}c@{\,~}@{\,~}|c@{\,~}@{\,~}c@{\,~}@{\,~}|c@{\,~}@{\,~}c@{\,~}@{\,~}|c@{\,~}@{\,~}c@{\,~}@{\,~}} 
				\toprule
				& \multicolumn{2}{c|}{4 Bands (NDVI)} & \multicolumn{2}{c|}{7 Bands (NDWI)} & \multicolumn{2}{c|}{8 Bands (NDRE)} &  \multicolumn{2}{c|}{10 Bands (NDBI)} &  \multicolumn{2}{c}{Average} \\ 
				\multirow{-2}{*}{\textbf{Method}}  & RMSE$\downarrow$ & MAE$\downarrow$  & RMSE$\downarrow$ & MAE$\downarrow$ & RMSE$\downarrow$	 & MAE$\downarrow$ & RMSE$\downarrow$  & MAE$\downarrow$ & RMSE$\downarrow$  & MAE$\downarrow$ \\
				\midrule
				DDPM & 0.502 & 0.441 & 0.341 & 0.266 & 0.884 & 0.871 & 0.593 & 0.514 & 0.648 & 0.596 \\
				ReFlow & 0.524 & 0.484 & \underline{0.317} & \underline{0.247} & 0.837 & 0.810 & 0.518 & 0.479 & 0.613 & 0.576 \\
				DDBM & 0.739 & 0.586 & 0.522 & 0.432 & 0.790 & 0.714 & 0.654 & 0.557 & 0.714 & 0.609 \\
				RDDM & 0.575 & 0.466 & 0.617 & 0.488 & 0.765 & 0.723 & 0.574 & 0.509 & 0.642 & 0.568 \\
				RDBM & \underline{0.428} & \underline{0.373} & 0.394 & 0.379 & \underline{0.422} & \underline{0.327} & \underline{0.464} & \underline{0.393} & \underline{0.434} & \underline{0.364} \\
				Ours(\ding{53}text) & \textbf{0.360} & \textbf{0.283} & \textbf{0.279} & \textbf{0.233} & \textbf{0.284} & \textbf{0.210} & \textbf{0.337} & \textbf{0.289} & \textbf{0.321} & \textbf{0.256} \\
				Ours(\ding{51}text) & 0.255 & 0.192 & 0.166 & 0.129 & 0.199 & 0.140 & 0.147 & 0.101 & 0.196 & 0.142 \\ 
				\bottomrule
		\end{tabular}}  
	\end{minipage}
	\vspace{-0.5em}
\end{table*}   

\textbf{Effects of CLA across different solvers.}
We evaluate CLA with Euler and DPM-Solver++(2M)~\cite{lu2025dpm} using identical random seeds, as shown in Tab.~\ref{tab:solver}. CLA consistently improves the average translation performance of both solvers, with more pronounced gains for Euler, demonstrating the effectiveness of progressively refining incomplete visual conditions during sampling. This difference can be attributed to their numerical characteristics. Euler directly updates the current state using timestep-wise velocity estimates, allowing CLA-induced condition refinement to immediately affect the transport trajectory, whereas DPM-Solver++(2M) incorporates historical model predictions in its multistep updates and is therefore less responsive to timestep-dependent corrections. Textual conditioning further provides complementary semantic guidance and improves translation fidelity across both solvers. Besides, Euler with CLA outperforms Euler with more sampling steps, under the 80 NFE budget, verifying that refining incomplete visual conditions is more effective than simply increasing the sampling resolution. Despite additional model evaluations, Euler with CLA and text achieves the best overall performance, offering a favorable efficiency–accuracy trade-off.

\begin{table}[t]
	\caption{Effect of CLA on different solvers for multimodal image translation tasks.}\label{tab:solver}
	\centering
	\vspace{-1.0em}
	\renewcommand\arraystretch{0.95}
	\resizebox{1.0\columnwidth}{!}{
		\begin{tabular}{ccc|cc|cc|cc|cc|cc|cc}
			\toprule
			\multicolumn{3}{c|}{\textbf{Method}} &  \multirow{2}{*}{\textbf{Step}} &  \multirow{2}{*}{\textbf{NFE}}  & \multicolumn{2}{c|}{PAN $\rightarrowtail$ MSI} & \multicolumn{2}{c|}{MSI $\rightarrowtail$ PAN} & \multicolumn{2}{c|}{SAR $\rightarrowtail$ MSI} & \multicolumn{2}{c|}{MSI $\rightarrowtail$ SAR} &  \multicolumn{2}{c}{Average} \\ 
			Solver & CLA & Text & & & PSNR$\uparrow$ & SSIM$\uparrow$ & PSNR$\uparrow$ & SSIM$\uparrow$ & PSNR$\uparrow$ & SSIM$\uparrow$ & PSNR$\uparrow$ & SSIM$\uparrow$ & PSNR$\uparrow$ & SSIM$\uparrow$ \\
			\midrule  
			Euler&\ding{53}&\ding{53} & 40 & 40 & 21.73 & 0.660 & 34.77 & 0.910 & 17.09 & 0.365 & 19.54 & 0.278 & 25.21 & 0.632\\	  			
			Euler&\ding{51}&\ding{53} & 40 & 80 & 22.07 & 0.662 & 35.17 & 0.914 & 17.36 & 0.373 & 20.32 & 0.291 & 25.63 & 0.639 \\	  
			Euler&\ding{53}&\ding{51} & 40 & 40 & 24.30 & 0.768 & 35.48 & 0.919 & 18.11 & 0.375 & 19.66 & 0.283 & 26.43 & 0.671 \\	  			
			\rowcolor{red!5}Euler&\ding{51}&\ding{51} & 40 & 80 & 25.16 & 0.786 & 36.10 & 0.925 & 18.39 & 0.382 & 20.50 & 0.302 & 27.10 & 0.683 \\	  
			Euler&\ding{53}&\ding{53} & 80 & 80 & 21.78 & 0.653 & 35.09 & 0.907 & 17.19 & 0.366 & 20.27 & 0.285 & 25.48 & 0.631 \\	  
			Euler&\ding{53}&\ding{51} & 80 & 80 & 24.58 & 0.776 & 35.96 & 0.925 & 18.21 & 0.369 & 20.39 & 0.301 & 26.83 & 0.678 \\	  
			DPMSolver&\ding{53}&\ding{53}& 40 & 40 & 21.53 & 0.633 & 34.61 & 0.911 & 16.57 & 0.324 & 20.13 & 0.281 & 25.10 & 0.619  \\
			DPMSolver&\ding{51}&\ding{53}& 40 & 80 & 21.73 & 0.636 & 34.90 & 0.913 & 16.77 & 0.331 & 20.20 & 0.284 & 25.31 & 0.622   \\
			DPMSolver&\ding{53}&\ding{51}& 40 & 40 & 24.30 & 0.757 & 35.28 & 0.921 & 16.85 & 0.343 & 20.34 & 0.290 & 26.26 & 0.664 \\
			DPMSolver&\ding{51}&\ding{51}& 40 & 80 & 24.31 & 0.758 & 35.55 & 0.922 & 17.92 & 0.350& 20.41 & 0.293 & 26.55 & 0.667   \\
			\bottomrule
	\end{tabular}}
\end{table}   

\section{Conclusion}
We present OmniRSI, a latent diffusion model for unified multimodal remote sensing image generation under heterogeneous and partially observed data. OmniRSI tightly couples a RS-VAE for shared multimodal representation, a fully adapted diffusion backbone for unified generation and translation, and a training-free cyclic latent alignment mechanism for progressively refining incomplete visual conditions during inference. Furthermore, we construct RSI-MM, a large-scale multimodal dataset spanning diverse sensors, resolutions, and modality configurations. Extensive experiments on reconstruction, text-to-image generation, and cross-modal translation demonstrate our superiority.
 
\subsection*{AI use statement}

In this work, we used generative AI tools for generating synthetic datasets (Appendix~\ref{sec:A}) and editing the manuscript to improve readability and language clarity. We have not used generative AI tools for literature searches or discovery, research ideation, drafting manuscript sections, or proving mathematical claims. We take full responsibility for the final content of this work, including text, claims or artifacts produced with the aid of generative AI.
 
%

\subsection*{Ethics and Reproducibility statement}

This work does not involve ethical concerns requiring specific consideration. Code and datasets will be provided in the \url{https://github.com/MiliLab/OmniRSI}.

\bibliography{main}
\bibliographystyle{iclr2027_conference}

\newpage

\appendix
\setcounter{equation}{0}
\renewcommand{\theequation}{\thesection.\arabic{equation}} 
\renewcommand{\thesection}{A}

\section{Textual Annotation Pipeline}\label{sec:A}
The datasets in Tab.~\ref{tab:RSI-MM} that lack textual annotations generally comprise paired RGB modalities. We exploit the availability of RGB images, either directly provided or extracted from the corresponding bands subsets, to establish a consistent visual basis for annotation. We feed these RGB visualizations to GPT-5.4-mini~\cite{singh2025openai} using a unified, evidence-oriented prompt, as shown in box~\ref{box:prompt}. The prompt covers eight aspects: scene category, primary objects, secondary elements, visual attributes, spatial layout, spatial relations, landuse Inference, and uncertainty. It prioritizes visible evidence and distinguishes inferred functions from direct observations. Each annotation forms a single coherent plain-text paragraph, as exemplified in Fig.~\ref{fig:annotation}. Annotation quality is further examined through random sampling and expert review.

\begin{tcolorbox}[
	title={A. The detailed textual descriptions of utilzed annotation prompt},
	label={box:prompt},  
]
\textbf{Prompt Template}
You are an expert in remote sensing image interpretation. Analyze the input remote sensing image in a structured and fine-grained manner.

Instructions: Focus on objective visual evidence first, then make cautious semantic inferences. Use professional remote sensing terminology when appropriate. Be as detailed as possible, but do not invent objects or relations that are not visually supported. If some regions are ambiguous due to resolution, shadow, cloud, occlusion, or image quality, explicitly state the uncertainty. Please output the analysis using the following fields:

1. Overall Scene Type: Identify the overall scene category or categories, such as urban area, industrial zone, port, airport, farmland, forest, river, coastal area, desert, mountain, or others.

2. Major Objects and Elements: List the main visible objects and land-use elements, such as buildings, roads, bridges, vehicles, ships, containers, runways, etc.

3. Minor or Secondary Elements: Describe smaller-scale or less dominant objects and scene elements that are still visually meaningful.

4. Visual Attributes: Describe the visible attributes of each object or region, including shape, size, color, texture, density, regularity, boundary clarity, orientation, etc.

5. Spatial Layout and Distribution: Describe how objects are distributed across the image. Consider patterns such as clustered, scattered, grid-like, linear, radial or irregular. Mention relative positions (center, edge, upper/lower/left/right regions if useful).

6. Spatial and Topological Relationships: Describe relations among objects, such as adjacent to, connected to, crossing, parallel to, surrounding, enclosed within, aligned along, separated by, bordering, intersecting, or extending toward.

7. Functional or Land-Use Interpretation: Based on visible evidence, infer the likely function of the scene or subregions, such as residential, commercial, industrial, logistics, transportation, agricultural, storage, recreational, or under construction.

8. Uncertainty: Explicitly identify objects, regions, or functional interpretations that remain uncertain, ambiguous, or tentative.

Output requirements: Provide a single coherent paragraph in plain text only, not bullet points, within 256 words. Ensure the description is comprehensive yet concise, avoid generic statements, and use natural sentences only.
\end{tcolorbox}

\renewcommand\thefigure{A1} 
\begin{figure*}[h] 
	\vspace{0pt}
	\centering
	\includegraphics[width=0.92\linewidth]{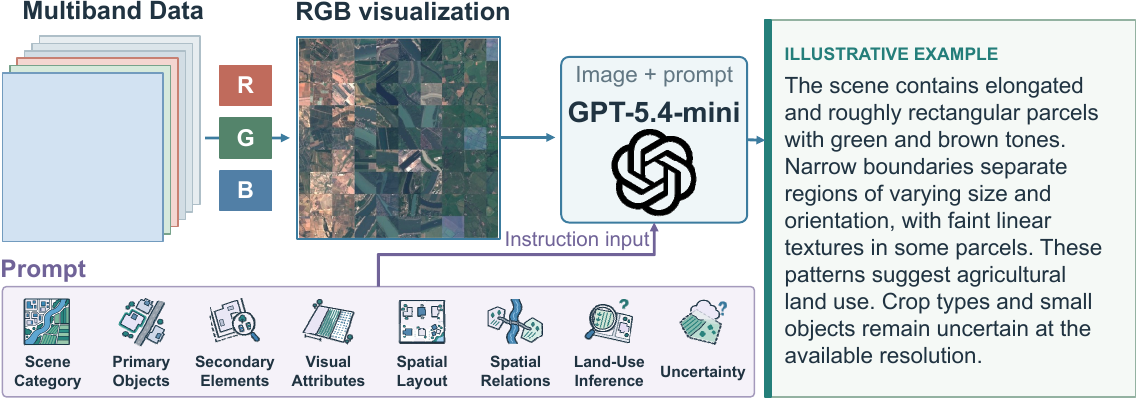}
	\vspace{-1.0em}
	\caption{Text annotation pipeline for RSI-MM. Representative visualizations are selected as inputs, followed by GPT-based caption generation with a fixed prompt template.}
	\label{fig:annotation} 
\end{figure*} 

\setcounter{equation}{0}
\renewcommand{\theequation}{\thesection.\arabic{equation}} 
\renewcommand{\thesection}{B}
\section{Details of Modality-aware Slots}\label{sec:B}
RSI-MM integrates heterogeneous observations from multiple satellite families, including Sentinel, Landsat, WorldView, GaoFen, and GeoEye, whose spectral configurations and modality combinations vary substantially across sensors. To organize heterogeneous remote sensing observations into a consistent input-output structure, we define $K\!=\!18$ modality-aware slots, which comprise one panchromatic slot, sixteen spectral slots spanning coastal to thermal infrared bands, and one SAR slot. Sensor-specific optical bands are mapped to the corresponding canonical slots according to their spectral characteristics, while PAN and SAR observations occupy dedicated slots. This design converts variable sensor configurations into a fixed channel ordering, enabling RS-VAE to jointly process heterogeneous and partially observed data. For each sample, slot availability is indicated by a binary mask $\mathbf{M}\!=\!\{\mathbf{m}_k\}_{k=1}^{K}$, where $\mathbf{m}_k\!=\!1$ denotes an observed slot and $\mathbf{m}_k\!=\!0$ denotes a missing one. Notably, these slots serve as \textbf{\emph{canonical band references}} rather than exact sensor specifications. The wavelength ranges and centers in Tab.~\ref{tab:slots} therefore define representative anchors for cross-sensor organization, and bands assigned to the same slot may still differ in spectral response, radiometric characteristics, and acquisition conditions.  Even observations from the same sensor and nominal band can vary substantially across seasons or imaging conditions, as exemplified by infrared imagery from the same sensor acquired in different seasons. Thus, the slot design establishes a common representation convention without assuming strict spectral or radiometric equivalence across sensors. Notably, we exclude hyperspectral imagery from the canonical slot space. Its densely sampled narrow bands would require substantially finer slot partitioning and greatly expand the unified input-output space. Moreover, large-scale hyperspectral data are less compatible with the current setting in terms of spatial resolution and scalable semantic annotation, making balanced joint learning more challenging. Extending the canonical space to hyperspectral observations remains an important direction for future work. 

\renewcommand\thetable{A1}
\begin{table*}[h]
	\centering
	\caption{Canonical slots and their sensor-specific band mappings. Wavelength ranges and centers are slot reference values. Bands assigned to the same slot need not have identical spectral responses.}\label{tab:slots}
	\vspace*{-0.6\floatsep}
	\renewcommand{\arraystretch}{1.1}
	\setlength{\tabcolsep}{4pt}
	\resizebox{\textwidth}{!}{%
		\begin{tabular}{clccccccc}
			\toprule
			\makecell{Slot} & Name
			& \makecell{Wavelength Range (nm)}
			& \makecell{Slot Wavelength Center (nm)}
			& \makecell{Sentinel-2A/B}
			& \makecell{Landsat 8/9}
			& \makecell{WorldView2/3}
			& \makecell{Landsat 7}
			& \makecell{GaoFen/GeoEye Series} \\
			\midrule
			0  & PAN               & Sensor-dependent    & ---   & ---   & B8  & PAN      & B8  & PAN   \\
			1  & Coastal           & 400--454            & 427   & B1    & B1  & Coastal  & --- & ---   \\
			2  & Blue              & 450--526            & 488   & B2    & B2  & Blue     & B1  & Blue  \\
			3  & Green             & 510--600            & 555   & B3    & B3  & Green    & B2  & Green \\
			4  & Yellow            & 585-–625            & 605   & ---   & --- & Yellow   & --- & ---   \\
			5  & Red               & 630--690            & 660   & B4    & B4  & Red      & B3  & Red   \\
			6  & RedEdge-1         & 695–-712            & 703.5 & B5    & --- & ---      & --- & ---   \\
			7  & RedEdge-2         & 705–-748            & 726.5 & B6    & --- & Red Edge & --- & ---   \\
			8  & RedEdge-3         & 769--793            & 781   & B7    & --- & ---      & --- & ---   \\
			9  & NIR-1             & 770--920            & 845   & B8    & --- & NIR1     & B4  & NIR1  \\
			10 & NIR-2             & 850-–1040           & 945   & B8A   & B5  & NIR2     & --- & ---   \\
			11 & Water Vapor       & 932--956            & 944   & B9    & --- & ---      & --- & ---   \\
			12 & Cirrus            & 1358–1392           & 1375  & B10   & B9  & ---      & --- & ---   \\
			13 & SWIR-1            & 1550--1750          & 1650  & B11   & B6  & ---      & B5  & ---   \\
			14 & SWIR-2            & 2090--2350          & 2220  & B12   & B7  & ---      & B7  & ---   \\
			15 & TIR-1             & 10600--11190        & 10895 & ---   & B10 & ---      & --- & ---   \\
			16 & TIR-2             & 11500--12510        & 12005 & ---   & B11 & ---      & --- & ---   \\
			17 & SAR               & ---                 & ---   & ---   & --- & ---      & --- & ---   \\
			\bottomrule
		\end{tabular}%
	}
	\smallskip
	\begin{minipage}{\textwidth}
		\textit{Notes.} Sentinel-2A/B band limits are approximated by the reported central wavelength plus or minus half the bandwidth; other bands use the adopted nominal ranges. The GaoFen mappings refer to GF-1/2/7 optical imaging cameras. Unused bands are omitted from this table.
	\end{minipage}	
\end{table*} 

\setcounter{equation}{0}
\renewcommand{\theequation}{\thesection.\arabic{equation}} 
\renewcommand{\thesection}{C}
\section{Datasets Split and Model Details}\label{sec:C}

\subsection{Datasets Details.}\label{sec:C-1}
RSI-MM integrates 16 public datasets to support multimodal image reconstruction, text-to-image generation, and image-to-image translation through the training of the VAE, SD3.5 Medium backbone, and ControlNet branch, respectively. All constituent datasets contribute their training samples to VAE optimization. Samples with image–text pairs, using either existing or newly supplemented captions, are used for text-to-image generation. Image-to-image translation is trained on samples with paired observations across the supported source–target modality configurations. Accordingly, FSARCap, Git-10M, Inria, RSCC, and RSICD are excluded from translation training because they contain only a single image modality.

For comprehensive evaluation, we first use source-provided metadata, including geographic coordinates and cropping information, to identify samples originating from the same larger scene and randomly retain one representative sample per group, thereby reducing scene redundancy. Furthermore, we exclude samples with registration-induced black borders or spatially homogeneous content, such as patches containing only open water or sand. Beyond the redundancy and quality filtering described above, we retain all remaining samples without dataset-level or additional scene-based selection, preserving scene diversity. According to the aforementioned protocol, we initially select 500 samples from each dataset for testing, except for OSCD and PSBench. OSCD is used exclusively for evaluation, whereas PSBench contributes 500 samples from each of its four spectral configurations, yielding 2,000 test samples. All remaining samples are used for training, with consistent partitions maintained across training stages. After filtering, the VAE is evaluated on 3,000 PAN/OPT/NIR samples, 3,500 SAR samples, and 500 MSI samples, following the spectral configuration of Any2Any for multimodal image reconstruction. The SD3.5 Medium backbone is evaluated on 7,000 image samples with associated textual descriptions for text-to-image generation, while the ControlNet branch is evaluated on 4,403 samples with paired observations for image-to-image translation. These counts refer to held-out image samples within each task, without separately counting multiple captions or translation directions associated with the same sample. 

Since the source datasets vary substantially in sample numbers and sizes, we adopt dataset-balanced sampling to prevent larger datasets from dominating optimization. For text-to-image training, the text condition is randomly dropped with a probability of 10\% to support classifier-free guidance. For ControlNet training, subsets of visual-condition channels are further randomly masked to improve robustness to varying and incomplete modality combinations.

\subsection{Model Details.}\label{sec:C-2}
\textbf{RS-VAE} aims to learn compact latent representations using an encoder $\mathcal{E}_{\theta}$ and a decoder $\mathcal{D}_{\theta}$. The encoder parameterizes a diagonal Gaussian posterior $q_{\theta}(\mathbf{Z}\mid\mathbf{X})
=\mathcal{N}(\boldsymbol{\mu}, \operatorname{diag}(\boldsymbol{\sigma}^{2}))$, where $\boldsymbol{\mu}$ and $\boldsymbol{\sigma}^{2}$ are the input-dependent posterior mean and variance. Latents are sampled via reparameterization:
\begin{equation}\label{eq:vae_reparameterization}
	\mathbf{Z}=\boldsymbol{\mu}+\boldsymbol{\sigma}\odot\boldsymbol{\epsilon},	\qquad \boldsymbol{\epsilon}\sim\mathcal{N}(0,\mathbf{I}),
\end{equation}
where the latent variable $\mathbf{Z}$ is then reconstructed as $\widehat{\mathbf{X}}=\mathcal{D}_{\theta}(\mathbf{Z})$. This process is optimized based on the evidence lower bound (ELBO):
\begin{equation}\label{eq:vae_elbo}
	\log p_{\theta}(\mathbf{X})	\geq{}\mathbb{E}_{q_{\theta}(\mathbf{Z}\mid\mathbf{X})}
	\left[\log p_{\theta}(\mathbf{X}\mid\mathbf{Z})\right]-	D_{\mathrm{KL}}\!\left(		q_{\theta}(\mathbf{Z}\mid\mathbf{X})\,\|\,\mathcal{N}(0,\mathbf{I})
	\right).
\end{equation}
The likelihood term promotes faithful reconstruction through $\mathcal{D}_{\theta}$, while the KL term regularizes the posterior toward an isotropic Gaussian prior. RS-VAE augments this formulation with the masked reconstruction and latent smoothness constraints, as described in Sec.~\ref{sec:method}.

After training RS-VAE, we freeze its parameters and encode the entire training set to estimate the global latent mean and variance across samples, channels, and spatial positions. We set the shift factor $b$ to the empirical mean and the scale factor $s$ to the reciprocal of the empirical standard deviation. These global statistics are distinct from the input-dependent posterior parameters $\boldsymbol{\mu}$ and $\boldsymbol{\sigma}^{2}$. During diffusion training and inference, encoder-produced latents are shifted and scaled before entering the diffusion model. The transformation is reversed before decoding:
\begin{equation}\label{eq:latent_normalization}
	\widetilde{\mathbf{Z}}=s(\mathbf{Z}-b),\qquad\widehat{\mathbf{X}}=\mathcal{D}_{\theta}\!\left(\widehat{\mathbf{Z}}/{s}+b\right),
\end{equation}
where $\mathbf{Z}$ is obtained from $\mathcal{E}_{\theta}$ and $\widehat{\mathbf{Z}}$ denotes the diffusion output in the normalized latent space. The estimated factors for RS-VAE are $s=0.8368$ and $b=-0.0881$, replacing the original SD3.5 VAE values of $s=1.5305$ and $b=0.0609$. These factors remain fixed throughout diffusion training and inference, standardizing the training latents to approximately zero global mean and unit global variance without requiring an exactly Gaussian distribution.

\textbf{SD3.5 Medium Backbone} comprises 24 multimodal diffusion transformer (MMDiT) blocks. Given a noisy latent $\mathbf{Z}_{t}\in\mathbb{R}^{16\times H\times W}$, it is partitioned into non-overlapping $2\times2$ patches through a convolutional projection, producing image tokens with $1,536$ dimensions. Spatial positional embeddings are center-cropped from a fixed $384\times384$ embedding grid, whose capacity is $147,456$ tokens; the actual token count depends on the input latent resolution. Text tokens are linearly projected from $4,096$ to $1,536$ dimensions. A $256$-dimensional sinusoidal timestep embedding and a $2,048$-dimensional pooled text representation are processed by separate two-layer MLPs with SiLU activation and summed to provide a $1,536$-dimensional modulation signal.

\renewcommand\thetable{A2}
\begin{table}[t]
	\centering
	\caption{Architecture configuration of the SD3.5 Medium backbone and ControlNet branch.}\label{tab:sd35_config}
	\vspace*{-0.6\floatsep}  
	\renewcommand{\arraystretch}{1.08}
	\begin{tabular}{ll}
		\toprule
		Configuration & Value \\
		\midrule
		\multicolumn{2}{l}{\textit{SD3.5 Medium backbone}} \\
		MMDiT blocks & 24 \\
		Latent input/output channels & 16 / 16 \\
		Latent patch size & $2\times2$ \\
		Hidden dimension & 1,536 \\
		Attention heads / head dimension & 24 / 64 \\
		Feed-forward expansion ratio & 4 \\
		Feed-forward hidden dimension & 6,144 \\
		Feed-forward activation & GELU (Tanh approximation) \\
		Query/key normalization & RMSNorm, $\epsilon=10^{-6}$ \\
		Block normalization & Non-affine LayerNorm, $\epsilon=10^{-6}$ \\
		Additional image self-attention & Blocks 0--12 \\
		Text-token projection & $4{,}096\rightarrow1{,}536$ \\
		Pooled-text embedding MLP
		& $2{,}048\rightarrow1{,}536\rightarrow1{,}536$ \\
		Timestep embedding MLP
		& $256\rightarrow1{,}536\rightarrow1{,}536$ \\
		Conditioning MLP activation & SiLU \\
		Positional embedding grid & $384\times384$ \\
		Positional embedding capacity & 147,456 positions \\
		Output projection per token & $1{,}536\rightarrow64$ \\
		\midrule
		\multicolumn{2}{l}{\textit{ControlNet branch}} \\
		Single-stream Transformer modules & 12 \\
		Noisy/condition latent channels & 16 / 16 \\
		Latent patch size & $2\times2$ \\
		Input fusion & Sum of separate patch embeddings \\
		Hidden dimension & 1,536 \\
		Attention heads / head dimension & 24 / 64 \\
		Feed-forward hidden dimension & 6,144 \\
		Feed-forward activation & GELU (tanh approximation) \\
		Block normalization & Non-affine LayerNorm, $\epsilon=10^{-6}$ \\
		Query/key normalization & None \\
		Explicit positional embeddings & None \\
		Joint image--text attention & None \\
		Pooled conditioning input dimension & 1,536 \\
		Pooled-text embedding MLP
		& $1{,}536\rightarrow1{,}536\rightarrow1{,}536$ \\
		Timestep embedding MLP
		& $256\rightarrow1{,}536\rightarrow1{,}536$ \\
		Conditioning MLP activation & SiLU \\
		Timestep input & Noise level $\times\,1{,}000$ \\
		Residual projections & 12 separate $1{,}536\rightarrow1{,}536$ layers \\
		Control strength & 1 \\
		Residual injection & After every backbone block \\
		Control-feature index at block $i$ & $\lfloor i/2\rfloor$, $i=0,\ldots,23$ \\
		\bottomrule
	\end{tabular}
\end{table}

Each MMDiT block maintains separate image and text parameters and performs joint attention over the concatenated image and text queries, keys, and values. Attention uses 24 heads with 64 dimensions per head and RMS normalization of queries and keys. The first 13 blocks additionally perform image-only self-attention. Adaptive LayerNorm and gated residual connections incorporate timestep and pooled-text conditioning, while feed-forward networks use a hidden dimension of 6,144 and GELU activation. The final block uses the text stream to construct attention inputs without further updating its representation. An adaptively normalized linear projection maps each image token to $2\times2\times16$ values, which are rearranged into a latent velocity prediction with the same spatial resolution and channel count as the input. Tab.~\ref{tab:sd35_config} summarizes the architecture configuration.

\textbf{ControlNet Branch} contains 12 independently parameterized single-stream Transformer modules, which share a block structure similar to MMDiT but differ in their attention interaction manner. The noisy latent and the observed-condition latent, both with 16 channels, are processed by separate $2\times2$ patch projections and summed to form a shared $1,536$-dimensional token sequence. Each module uses 24-head self-attention with a head dimension of 64, adaptive LayerNorm, gated residual connections, and a feed-forward network with GELU activation. The control branch does not use explicit positional embeddings, query/key normalization, or joint attention with text tokens.
 
Control conditioning combines a timestep embedding, computed from the noise level multiplied by a certain factor, with a pooled-text embedding. The original $2,048$-dimensional pooled text representation is first mapped to $1,536$ dimensions by the backbone's text embedder and then processed by a separate MLP in ControlNet. Both conditioning MLPs use SiLU activation. Each control-module output is transformed by a dedicated linear projection, producing 12 residual features. With the control strength fixed to 1, these features are injected into the backbone image stream as:
\begin{equation}\label{eq:controlnet_injection}
	\mathbf{H}_{i}\leftarrow \mathbf{H}_{i}+	\mathbf{h}^{c}_{\lfloor i/2 \rfloor },
	\qquad i=0,\ldots,23,
\end{equation} 
where $\mathbf{H}_{i}$ denotes the image-token representation after the $i$-th backbone block and $\mathbf{h}^{c}_{j}$ is the $j$-th projected control feature. Thus, each control feature is added after each of two consecutive backbone blocks. Notably, the official ControlNet architecture and parameters corresponding to SD3.5 Medium have not been open-sourced. The aforementioned process was modified based on the ControlNet branch of SD3.5 Large; however, the SD3.5 Large model is difficult for us to train.

\setcounter{equation}{0}
\renewcommand{\theequation}{\thesection.\arabic{equation}} 
\renewcommand{\thesection}{D}
\section{More Experiment Results}\label{sec:D}

\renewcommand\thefigure{A2} 
\begin{figure*}[h] 
	\vspace{0pt}
	\centering
	\includegraphics[width=1.0\linewidth]{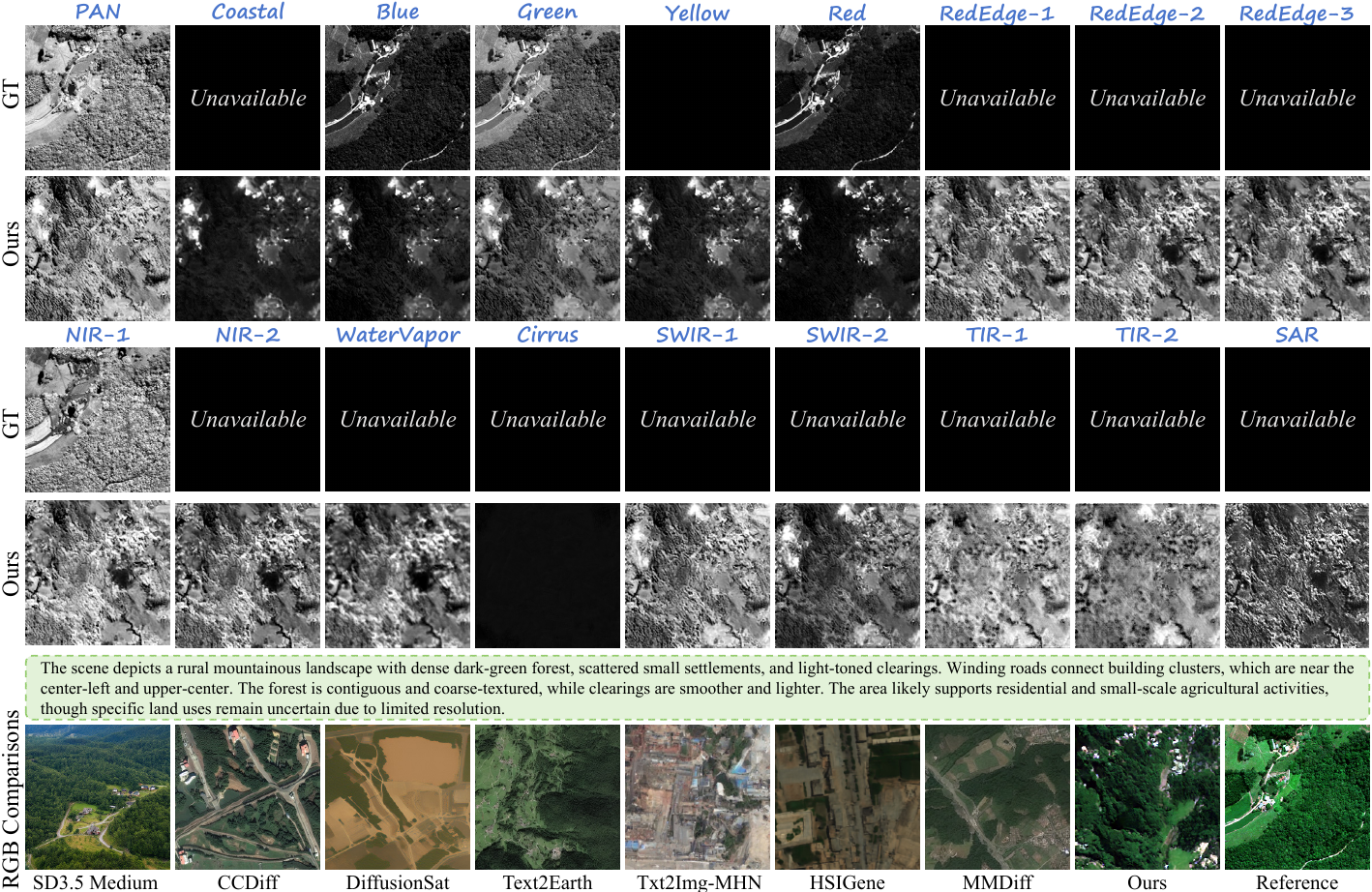}
	\vspace{-1.0em}
	\caption{Qualitative results of text-to-image Generation.}
	\label{fig:t2ifull} 
\end{figure*}

\renewcommand\thefigure{A3} 
\begin{figure*}[h] 
	\vspace{0pt}
	\centering
	\includegraphics[width=1.0\linewidth]{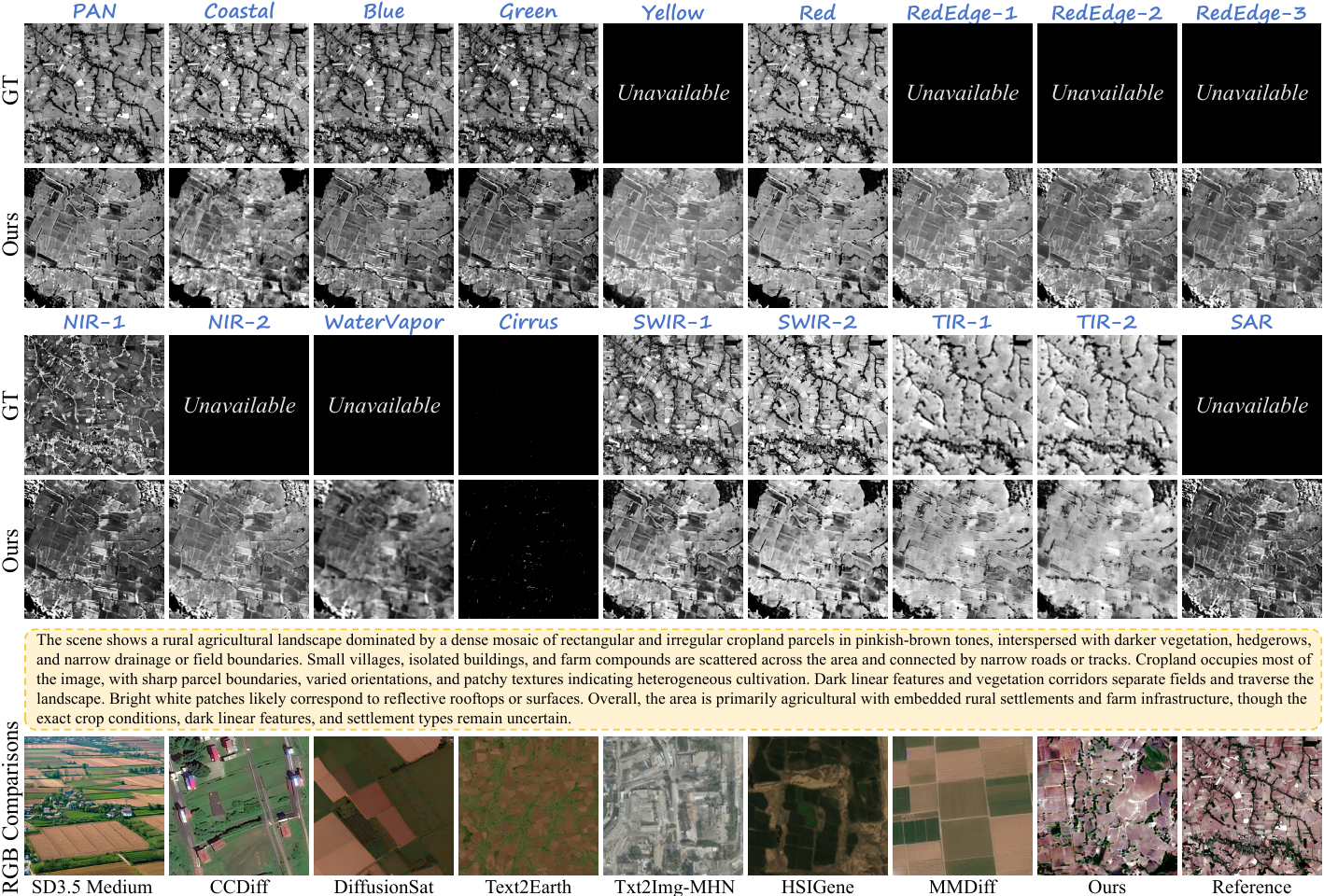}
	\vspace{-1.0em}
	\caption{Qualitative results of text-to-image Generation.}
	\label{fig:t2ifull2} 
\end{figure*}
 
\renewcommand\thefigure{A4} 
\begin{figure*}[htp] 
	\vspace{0pt}
	\centering
	\includegraphics[width=1.0\linewidth]{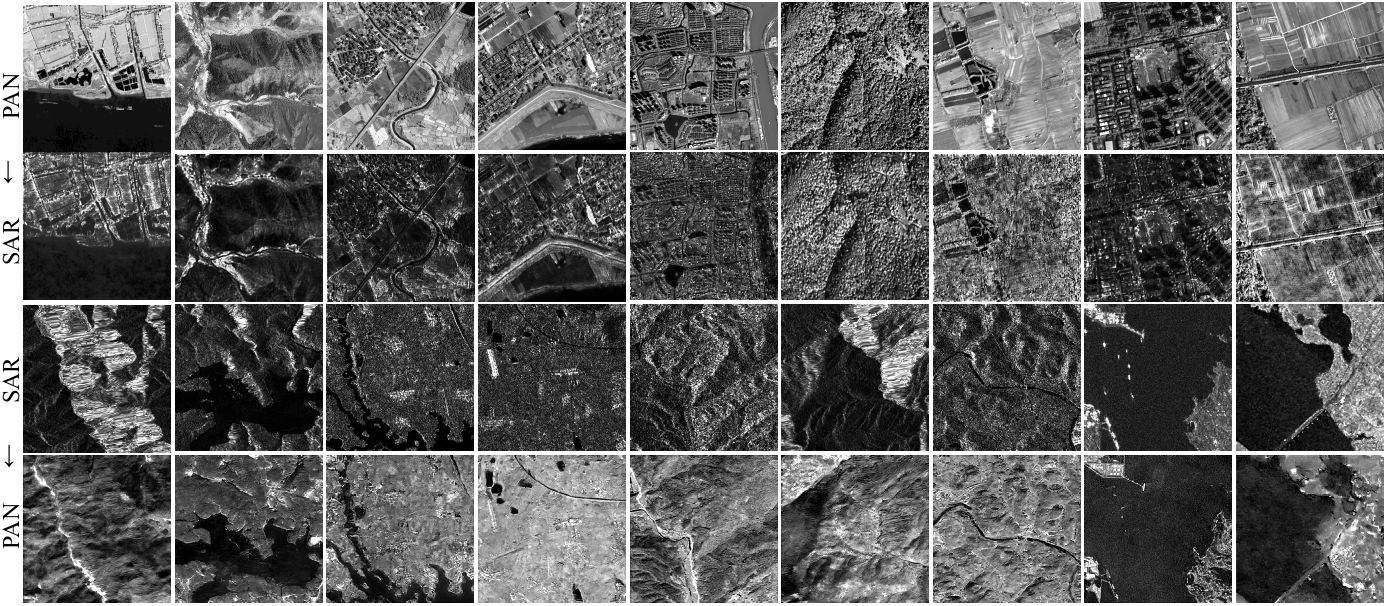}
	\vspace{-1.0em}
	\caption{Zero-shot translation from PAN to SAR (top) and from SAR to PAN (bottom).}
	\label{fig:pan2sar} 
\end{figure*}

\subsection{Multimodal Reconstruction}\label{sec:D-0}
We evaluate reconstruction under varying band availability on a selected subset of the EuroSAT test set, whose Sentinel-2 imagery contains 13 spectral bands. We progressively mask valid bands in a fixed order and evaluate self-reconstruction (Self-Rec) over the remaining observed bands and full reconstruction (Full-Rec) over all original bands, including those withheld from the input. As shown in Tab.~\ref{tab:rec}, RS-VAE maintains good reconstruction quality even with a single observed band and recovers information in withheld bands, suggesting that it captures useful spectral dependencies. As more bands become available, Full-Rec improves, while Self-Rec exhibits an overall upward trend with fluctuations as the evaluated band subset changes. The two settings coincide when all bands are observed. These results illustrate RS-VAE's ability to reconstruct sparse observations and exploit complementary spectral information for full-band recovery on the evaluated subset.

\renewcommand\thetable{A3}
\begin{table*}[h]
	\centering
	\caption{Reconstruction performance of RS-VAE with varying numbers of bands on EuroSAT (50 samples). Self-Rec evaluates the remaining observed bands, whereas Full-Rec evaluates all 13 original bands. Higher PSNR (dB) and SSIM indicate better reconstruction.}\label{tab:rec}
	\vspace*{-0.6\floatsep}
	\renewcommand{\arraystretch}{1.1}
	\setlength{\tabcolsep}{4pt}
	\resizebox{\textwidth}{!}{%
		\begin{tabular}{cccccccccccccc}
			\toprule
			Valid Band & 1 & 2 & 3 & 4 & 5 & 6 & 7 & 8 & 9 & 10 & 11 & 12 & 13 \\
			\midrule
			Self-Rec PSNR & 32.66 & 33.45 & 34.61 & 35.61 & 35.86& 37.95 & 38.95 & 38.97 & 39.02 & 41.44 & 43.01 & 43.95 & 45.58 \\
			Self-Rec SSIM & 0.930 & 0.940 & 0.954 & 0.959 & 0.966  & 0.969 & 0.974 & 0.977 & 0.979 & 0.986 & 0.988 & 0.990 & 0.991 \\
			Full-Rec PSNR & 29.75 & 32.59 & 33.42 & 33.45 & 33.58 & 36.04 & 36.82 & 37.03 & 37.20 & 41.15 & 42.68 & 43.78 & 45.58 \\
			Full-Rec SSIM & 0.904 & 0.929 & 0.935 & 0.936 & 0.936 & 0.961 & 0.968 & 0.972 & 0.975 & 0.987 & 0.989 & 0.990 & 0.991 \\
			\bottomrule
		\end{tabular}%
	}
\end{table*}

\subsection{Text-to-image Generation}\label{sec:D-1}
Existing remote sensing generation models predominantly target individual modalities or fixed band configurations. In contrast, OmniRSI models PAN, MSI, and SAR observations within a unified output space. Figs.~\ref{fig:t2ifull} and~\ref{fig:t2ifull2} complement the RGB comparisons by visualizing all modality slots. Conditioned on a single text prompt, OmniRSI generates spatially consistent structures across heterogeneous spectral and sensing modalities, including those absent from the corresponding reference observation. Black GT panels denote unavailable modalities. For visualization, each band is independently normalized by its maximum value. These results qualitatively demonstrate the capability of OmniRSI for unified full-modality remote sensing generation.

\begin{figure*}[t]
	\centering  
	 
	\vspace{0pt}
	\renewcommand\thefigure{A5} 
	\begin{minipage}{\textwidth}
		\centering
		\includegraphics[width=0.99\linewidth]{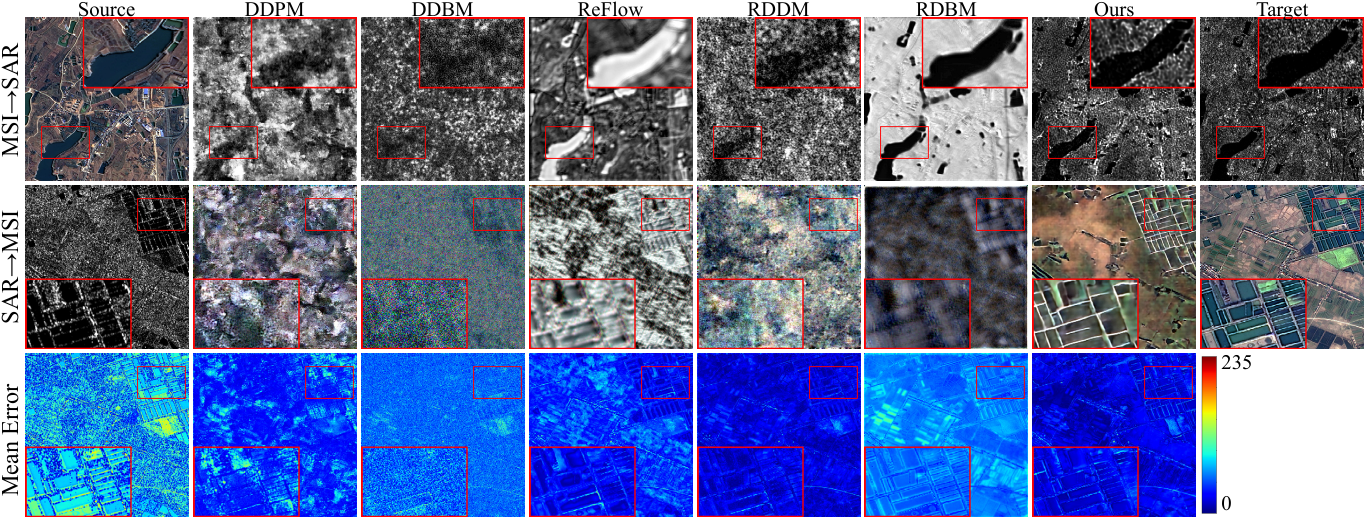}
		\caption{Qualitative results of image-to-image translation. Displayed in RGB.}
		\label{fig:i2i2}
	\end{minipage} 
	
	\vspace{0.0\floatsep}
	
	\renewcommand\thefigure{A6} 
	\begin{minipage}{\textwidth}
		\centering
		\includegraphics[width=1.0\linewidth]{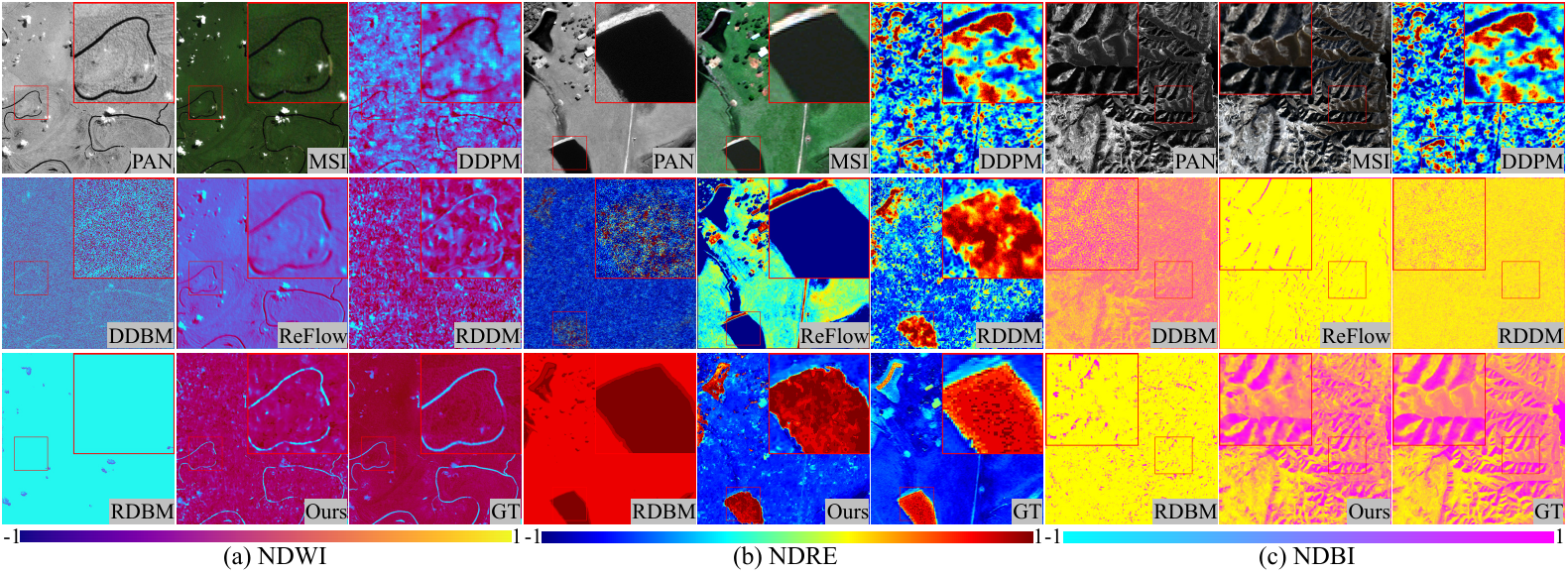}
		\caption{Visualizations of (a) NDWI, (b) NDRE and (c) NDBI (PAN$\rightarrow$MSI).}
		\label{fig:ndxi} 
	\end{minipage} 
	
	\vspace{0.0\floatsep}
	
	\renewcommand\thefigure{A7} 
	\begin{minipage}{\textwidth}
		\vspace{0pt}
		\centering
		\includegraphics[width=1.0\linewidth]{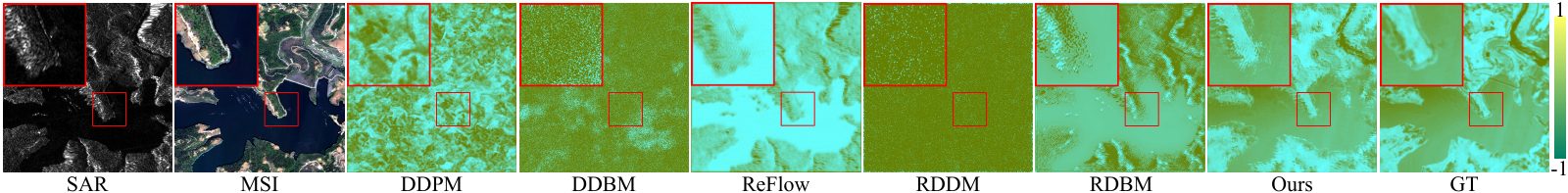}
		\caption{Visualization comparison of NDVI (SAR$\rightarrow$MSI).}
		\label{fig:sarndvi} 
	\end{minipage} 
\end{figure*}

\subsection{Image-to-image Translation}\label{sec:D-2}
OmniRSI supports unified image-to-image translation among three representative remote sensing modality groups, including PAN, MSI, and SAR. The main paper reports quantitative results and representative visualizations for PAN$\leftrightarrow$MSI translation. Here, we provide additional qualitative comparisons to further examine the cross-modal generation capability of OmniRSI. As shown in Fig.~\ref{fig:i2i2}, OmniRSI produces visually consistent SAR$\leftrightarrow$MSI translations while preserving major scene structures and modality-specific appearance characteristics. Since paired PAN--SAR observations are unavailable in RSI-MM, PAN$\leftrightarrow$SAR translation is evaluated in a zero-shot setting without task-specific paired supervision. As shown in Fig.~\ref{fig:pan2sar}, OmniRSI remains capable of transferring scene structures across previously unpaired modalities and producing plausible target-modality observations. Overall, OmniRSI exhibits consistently strong cross-modal translation performance across both paired and zero-shot modality settings, highlighting its generalization capacity.

To examine the motivation for CLA, we evaluate MSI$\rightarrow$PAN translation on PSBench (Landsat9) by progressively masking valid MSI bands in a fixed order. Tab.~\ref{tab:cla_support} shows that richer spectral conditions generally improve PAN reconstruction, although performance varies with the retained band subset. Increasing the available bands from one to ten raises both PSNR and SSIM to an extent. These results motivate enriching incomplete visual conditions during sampling, while the solver ablation in the main paper provides direct evidence of CLA's effectiveness.

\renewcommand\thetable{A4}
\begin{table*}[h]
	\centering
	\caption{PAN translation performance with varying numbers of observed MSI bands on PSBench (50 samples, MSI$\rightarrow$PAN). Bands are progressively masked in a fixed order.}\label{tab:cla_support}
	\vspace*{-0.6\floatsep}
	\renewcommand{\arraystretch}{1.0}
	\setlength{\tabcolsep}{4pt}
	\resizebox{0.85\textwidth}{!}{%
		\begin{tabular}{ccccccccccc}
			\toprule
			Valid Band & 1 & 2 & 3 & 4 & 5 & 6 & 7 & 8 & 9 & 10 \\
			\midrule
			PSNR & 26.22 & 30.60 & 33.32 & 33.76 & 34.06 & 37.36 & 37.63 & 39.18 & 40.92 & 41.67 \\
			SSIM & 0.671 & 0.929 & 0.937 & 0.946 & 0.951 & 0.965 & 0.967 & 0.971 & 0.974 & 0.975 \\
			\bottomrule
		\end{tabular}%
	} 
\end{table*} 

\subsection{NDXI Application}\label{sec:D-3}
 
To further assess the spectral fidelity of translated MSI observations, we evaluate normalized difference indices (NDXIs) on both PAN$\rightarrow$MSI and SAR$\rightarrow$MSI tasks. Since the available spectral bands vary across MSI configurations, we compute the corresponding index according to band availability: normalized difference vegetation index (NDVI)~\cite{tucker1979red} for four-band MSI, water index (NDWI)~\cite{mcfeeters1996use} for seven-band MSI, red edge index (NDRE)~\cite{barnes2000coincident} for eight-band MSI, and building index (NDBI)~\cite{zha2003use} for ten-band MSI. Their definitions are given below:
\begin{align} 
	\text{NDVI} &= \frac{NIR - R}{NIR + R}, \\ 
	\text{NDWI} &= \frac{G - NIR}{G + NIR}, \\
	\text{NDRE} &= \frac{NIR - RE}{NIR + RE}, \\
	\text{NDBI} &= \frac{SWIR - NIR}{SWIR + NIR}.
\end{align}
In accordance with these definitions, the normalized difference indices are confined to the range of $[-1, 1]$, which can serve as the spectral accuracy indicators of the translation process to a certain extent. The quantitative results over these four indices are presented in Tab.~\ref{tab:ndxi}. Clearly, our method achieves the best values across all these objective metrics, namely root mean squared error (RMSE), and mean absolute error (MAE). Moreover, we present the visualizations in Fig.~\ref{fig:ndxi} and Fig.~\ref{fig:sarndvi}. Particularly, all of our index maps exhibit the closest resemblance to the reference, while the other methods suffer from intensity distortions.

\subsection{Efficiency Comparison}\label{sec:D-4}
Tab~\ref{tab:effiency} compares peak GPU memory, inference latency, and throughput across resolutions. For image reconstruction, RS-VAE achieves substantially improved computational efficiency over Any2any, reducing peak memory at both spatial resolutions while attaining higher throughput. This indicates favorable scalability of the proposed VAE for high-resolution remote sensing reconstruction. For text-to-image generation, OmniRSI introduces moderate additional overhead over the SD3.5 Medium backbone owing to the incorporated multimodal conditioning modules. Nevertheless, its peak memory increases only marginally when the resolution is doubled, indicating stable memory scaling. The inference latency increases correspondingly, reflecting the additional conditional processing required by the unified multimodal framework. Multimodal image translation incurs the largest computational cost, as CLA performs additional conditional denoiser evaluations and repeatedly reconstructs and re-encodes intermediate multimodal observations during sampling. Consequently, OmniRSI exhibits higher inference latency at both spatial resolutions. Despite this increased latency, its peak memory remains nearly unchanged across resolutions, suggesting that the major overhead of CLA arises from additional computation rather than memory consumption.

\renewcommand\thetable{A5}
\begin{table}[htbp]
	\centering 
	\caption{Efficiency comparisons among different methods.}\label{tab:effiency}
	\vspace{-0.5\floatsep}
	\resizebox{0.75\textwidth}{!}{%
	\begin{tabular}{l*{6}{c}}
		\toprule
		Resolution & \multicolumn{3}{c}{256$\times$256} & \multicolumn{3}{c}{512$\times$512} \\ 
		\cmidrule(lr){2-4}\cmidrule(lr){5-7} 
		Method & Mem.(G) & Time(s) & FPS & Mem.(G) & Time(s) & FPS \\ 
		\midrule
		\multicolumn{7}{l}{\textbf{VAE Reconstruction}}\\
		\hline
		Any2any(MSI)   &  2.683 & 0.027  & 37.037 & 40.530 & 0.279  & 3.579  \\ 
		RS-VAE(Ours)   & 0.359 & 0.018 & 54.969 & 0.952 & 0.066 & 15.110 \\ 
		\hline
		\multicolumn{7}{l}{\textbf{Text-to-image Generation}}\\
		\hline
		 CC-Diff       & 7.961 & 2.974 & 0.336  & 9.832 & 5.805 & 0.172 \\ 
		 Txt2Img-MHN   & 0.581 & 1.527  & 0.655 & 0.980 & 6.591 & 0.152 \\ 
		 HSIGene       & 3.896 & 7.396 & 0.135 & 19.044 & 23.852 & 0.042 \\ 
		 Text2Earth    & 2.585 & 1.419 & 0.705  & 3.030 & 1.431 & 0.699\\ 
		 DiffusionSat  & 2.612 & 0.558 & 1.794 & 3.051 & 0.632 & 1.582 \\ 
		 MMDiff        & 2.172 & 1.503 & 0.665  & 2.617 & 1.541 & 0.649 \\ 
		 SD3.5 Medium  & 15.362 & 1.392 & 0.719 & 15.808 & 1.639 & 0.610 \\ 
		 OmniRSI(Ours) & 16.467 & 2.504  & 0.399 & 17.066 & 3.473 & 0.288 \\ 
		 \hline
		\multicolumn{7}{l}{\textbf{Multimodal Image Translation}}\\
		\hline
		DDPM    & 0.384 & 0.416  & 2.401 & 1.295 & 1.679  & 0.596 \\ 
		ReFlow  & 0.367 & 0.206  & 4.862 & 1.226 & 0.779 & 1.283 \\ 
		DDBM  	& 0.323 & 3.024  & 0.331 & 0.915 & 4.706 & 0.212 \\ 
		RDDM    & 0.378 & 0.850 & 1.176  & 1.331 & 3.285 & 0.304 \\ 
		RDBM    & 0.637 & 0.151 & 6.627 & 1.689 & 0.590 & 1.695 \\ 
		OmniRSI(Ours) & 16.482 & 5.611 & 0.178 & 17.111 & 9.462 & 0.106 \\ 
		\bottomrule
	\end{tabular}
}
\end{table}
 
\setcounter{equation}{0}
\renewcommand{\theequation}{\thesection.\arabic{equation}} 
\renewcommand{\thesection}{E} 
\section{Datasheets Details}\label{sec:E}

\subsection{Motivation}

\begin{enumerate}
\item \textit{``For what purpose was the dataset created?''}

\textcolor{BurntOrange}{\textbf{A:}} 
RSI-MM was created to support unified multimodal remote sensing image generation from heterogeneous and partially observed Earth observation data. Existing resources differ substantially in sensor characteristics, spatial resolution, spectral coverage, modality pairing, and text availability, making it difficult to learn and evaluate shared generative representations. RSI-MM brings together RGB, multispectral (MSI), synthetic aperture radar (SAR), and panchromatic (PAN) imagery with scene descriptions and partially paired observations. It supports multimodal reconstruction, text-to-image generation, and cross-modal translation, and provides the training and evaluation data for OmniRSI.

\item \textit{``Who created the dataset (\textit{e.g.}, which team, research group) and on behalf of which entity?''}

\textcolor{BurntOrange}{\textbf{A:}} RSI-MM was constructed by the authors of this work. Their name and institutional identities are omitted for anonymous review. The underlying imagery originates from the contributors and providers of the constituent public datasets.

\item \textit{``Who funded the creation of the dataset?''}

\textcolor{BurntOrange}{\textbf{A:}}
	The dataset creation was funded by the affiliations of the authors involved in this work.
\end{enumerate}

\subsection{Composition}
Most of the questions in this section are intended to provide dataset consumers with the information they need to make informed decisions about using the dataset for their chosen tasks. Some of the questions are designed to elicit information about compliance with the EU’s General Data Protection Regulation (GDPR) or comparable regulations in other jurisdictions. Questions that apply only to datasets that relate to people are grouped together at the end of the section. We recommend taking a broad interpretation of whether a dataset relates to people. For example, any dataset containing text that was written by people relates to people.
\begin{enumerate}
	\item \textit{``What do the instances that comprise our datasets represent (\textit{e.g.}, documents, photos, people, countries)?''}
	
	\textcolor{BurntOrange}{\textbf{A:}} Each instance represents a remote sensing image or image patch depicting an Earth surface scene, together with its associated textual descriptions and available modality observations. Depending on the source, an instance may include RGB, multispectral (MSI), synthetic aperture radar (SAR), or panchromatic (PAN) data, or a paired subset of these modalities. RGB composites derived from MSI bands are visual representations of the corresponding observations and do not constitute additional independent acquisitions.
	 
	\item \textit{``How many instances are there in total (of each type, if appropriate)?''}
	
	\textcolor{BurntOrange}{\textbf{A:}} RSI-MM contains 11,518,729 image samples aggregated from 16 public datasets and 11,864,029 text descriptions. The source-wise counts reported in the manuscript sum to 11,518,729: Git-10M (10,452,649), EuroSAT (27,000), OSCD (48), M3M-CR (63,000), OpenEarthMap-SAR (5,033), PIE-RGB-SAR (4,865), WHU-OPT-SAR (4,900), SARLO (119,566), RSCC (62,351), RSICD (10,921), PSBench (449,936), FSARCap (14,480), GID (25,350), Inria (1,800), M4-SAR (112,174), and ChatEarthNet (164,656). These are the listed source counts, not a verified count of globally unique scenes after cross-source deduplication. Details can be found in the Tab~\ref{tab:RSI-MM}.
	
	\item \textit{``Does the dataset contain all possible instances or is it a sample (not necessarily random) of instances from a larger set?''}
	
	\textcolor{BurntOrange}{\textbf{A:}} RSI-MM aggregates 16 public remote sensing datasets and is neither an exhaustive collection nor a uniformly random sample of global Earth observations. Its geographic coverage, scene distribution, spatial resolutions, and modality combinations inherit the selection characteristics of the constituent sources.	  Existing text annotations are used where available, and descriptions are generated for sources lacking text. The textual annotations are partially created by RSI-MM. Scene redundancy and image-quality filtering are addressed by the protocol in Appendix~\ref{sec:C-1}.
	
	\item \textit{``Is there a label or target associated with each instance?''}
	
	\textcolor{BurntOrange}{\textbf{A:}}Yes. Textual scene descriptions provide semantic conditioning, while available image bands serve as reconstruction or generation targets. Paired source and target observations support supervised cross-modal translation where such pairs are available. A binary availability mask identifies observed slots; missing modalities are not measured ground truth. 
	
	\item \textit{``Is any information missing from individual instances?''}
	
	\textcolor{BurntOrange}{\textbf{A:}} Yes. Partial modality and band availability is an intentional characteristic of RSI-MM. Observations are organized into 18 canonical slots: one PAN slot, 16 spectral slots spanning coastal to thermal infrared bands, and one SAR slot. A binary mask distinguishes available measurements from absent observations. Not every sample contains all slots, and paired PAN--SAR observations are unavailable in RSI-MM. Missing entries must not be interpreted as physical zero-valued measurements.
	
	\item \textit{``Are relationships between individual instances made explicit (\textit{e.g.}, users’ movie ratings, social network links)?''}
	
	\textcolor{BurntOrange}{\textbf{A:}} Yes. Relationships are represented through the association of captions with their images and the paired observations supplied by the constituent datasets. The modality-slot mapping identifies how available bands enter the shared representation. However, this organization does not establish pairing between arbitrary images from different sources, guarantee exact temporal correspondence, or imply that bands assigned to the same slot have identical sensor responses.  
	
	\item \textit{``Are there recommended data splits (\textit{e.g.}, training, development/validation, testing)?''}
	
	\textcolor{BurntOrange}{\textbf{A:}} 
	Yes. The manuscript reports randomly selecting partial samples from each constituent dataset for testing. An exact training/validation/test split protocol is provided in Appendix~\ref{sec:C-1}.
	
	\item \textit{``Is the dataset self-contained, or does it link to or otherwise rely on external resources (\textit{e.g.}, websites, tweets, other datasets)?''}
	
	\textcolor{BurntOrange}{\textbf{A:}} RSI-MM relies on 16 external public datasets for source imagery and, where available, existing captions. Newly generated captions supplement sources without text. It will be released with bundle images, annotations, source metadata, and usage documentation on platforms such as GitHub and Hugging Face, subject to the licenses and terms of the underlying source datasets.
	
	\item \textit{``Does the dataset contain data that might be considered confidential (\textit{e.g.}, data that is protected by legal privilege or by doctor–patient confidentiality, data that includes the content of individuals’ non-public communications)?''}
	
	\textcolor{BurntOrange}{\textbf{A:}} No. To the best of our knowledge, RSI-MM does not contain confidential records, private communications, or medical, financial, or legal records. The images are derived from publicly released remote sensing datasets, and annotations are constructed for benchmark evaluation.
	
	\item \textit{``Does the dataset contain data that, if viewed directly, might be offensive, insulting, threatening, or might otherwise cause anxiety?''}
	
	\textcolor{BurntOrange}{\textbf{A:}} No. RSI-MM collection is intended to describe Earth surface scenes and is not designed to include offensive or graphic content.
\end{enumerate}

\subsection{Collection Process}
In addition to the goals outlined in the previous section, the questions in this section are designed to elicit information that may help researchers and practitioners create alternative datasets with similar characteristics. Again, questions that apply only to datasets that relate to people are grouped together at the end of the section.
\begin{enumerate}
	\item \textit{``How was the data associated with each instance acquired?''}
	
	\textcolor{BurntOrange}{\textbf{A:}} 
	The images and available text annotations are obtained from 16 existing public datasets: Git-10M, EuroSAT, OSCD, M3M-CR, OpenEarthMap-SAR, PIE-RGB-SAR, WHU-OPT-SAR, SARLO, RSCC, RSICD, PSBench, FSARCap, GID, Inria, M4-SAR, and ChatEarthNet. For sources without textual annotations, representative visualizations are used to generate scene descriptions. RSI-MM construction is described as dataset aggregation and annotation, without a new image acquisition campaign. Please see Appendix~\ref{sec:A} for details.
	
	\item \textit{``What mechanisms or procedures were used to collect the data (\textit{e.g.}, hardware apparatuses or sensors, manual human curation, software programs, software APIs)?''}
	
	\textcolor{BurntOrange}{\textbf{A:}} 
	The source datasets provide observations from heterogeneous optical and radar sensors, including Sentinel, Gaofen, PlanetScope, Umbra, and other platforms listed in the manuscript. For missing text annotations, RGB images supplied by the sources or extracted from suitable band subsets were provided to GPT-5.4-mini using a unified, evidence-oriented prompt. The prompt covers scene category, primary objects, secondary elements, visual attributes, spatial layout, spatial relations, land-use inference, and uncertainty. Each generated caption is a single coherent plain-text paragraph. The prompt prioritizes visible evidence and distinguishes inferred functions from direct observations. A systematic human-verification procedure, combining random sampling with expert review, is conducted to ensure the correctness of the annotations.
	 
	\item \textit{``If the dataset is a sample from a larger set, what was the sampling strategy (\textit{e.g.}, deterministic, probabilistic with specific sampling probabilities)?''} 
	
	\textcolor{BurntOrange}{\textbf{A:}}RSI-MM is an aggregation of 16 public datasets rather than a sample from a larger set. See Sec.~\ref{sec:method} and Appendix~\ref{sec:C-1} for details.
\end{enumerate}

\subsection{Preprocessing, Cleaning, and Labeling}
The questions in this section are intended to provide dataset
consumers with the information they need to determine whether the “raw” data has been processed in ways that are compatible with their chosen tasks. For example, text that has been converted into a ``bag-of-words" is not suitable for tasks involving word order.
\begin{enumerate}
	\item \textit{``Was any preprocessing/cleaning/labeling of the data done (\textit{e.g.}, discretization or bucketing, tokenization, part-of-speech tagging, SIFT feature extraction, removal of instances, processing of missing values)?''}
	
	\textcolor{BurntOrange}{\textbf{A:}} 
	Yes. We extract RGB composites for caption generation, supplement missing textual annotations, and map available observations into modality-aware slots using sensor-specific band mappings. Binary availability masks distinguish missing entries from observed measurements. Optical bands are mapped according to their spectral characteristics, while PAN and SAR occupy dedicated slots. This mapping defines a common channel ordering without establishing radiometric equivalence or spectral harmonization across sensors. The data preparation protocol also includes scene redundancy reduction and filtering of registration-induced black borders and spatially homogeneous samples. See Appendix~\ref{sec:A} for annotation, Appendix~\ref{sec:C-1} for filtering and splits, and Tab.~\ref{tab:slots} for band mappings.
	
	\item \textit{``Was the `raw' data saved in addition to the preprocessed/cleaned/labeled data (\textit{e.g.}, to support unanticipated future uses)?''} 
	
	\textcolor{BurntOrange}{\textbf{A:}}  Yes. We retained the raw source data in addition to the preprocessed and annotated data. Access to copies distributed by the original source providers remains subject to their availability and terms.
	
	\item \textit{``Is the software that was used to preprocess/clean/label the data available?''} 
	
	\textcolor{BurntOrange}{\textbf{A:}} Yes. The preprocessing and evaluation scripts are planned for public release with the camera-ready version of the paper.
\end{enumerate}

\subsection{Uses}
The questions in this section are intended to encourage dataset creators to reflect on tasks for which the dataset should and should not be used. By explicitly highlighting these tasks, dataset creators can help dataset consumers make informed decisions, thereby avoiding potential risks or harms.
\begin{enumerate}
	\item \textit{``Has the dataset been used for any tasks already?''} 
	
	\textcolor{BurntOrange}{\textbf{A:}} 
	Yes. RSI-MM is used to train and evaluate OmniRSI for multimodal image reconstruction, text-to-image generation, and multimodal image translation. The study evaluates PAN--MSI and SAR--MSI translation in both directions and presents zero-shot PAN--SAR translation where paired training observations are unavailable. It also assesses spectral fidelity using indices derived from the available bands.
	
	\item \textit{``Is there a repository that links to any or all papers or systems that use the dataset?''} 
	
	\textcolor{BurntOrange}{\textbf{A:}} Yes. we will provide the links on the project GitHub and Hugging Face repositories.
	
	\item \textit{``What (other) tasks could the dataset be used for?''}
	
	\textcolor{BurntOrange}{\textbf{A:}} 
	RSI-MM may support research on image restoration, multi-source image fusion, image captioning, and multimodal representation learning. It may also support pretraining or data augmentation for downstream detection and segmentation, provided that the task-specific annotations required for training and evaluation are available or separately obtained.
	
	\item \textit{``Is there anything about the composition of the dataset or the way it was collected and preprocessed/cleaned/labeled that might impact future uses?''} 
	
	\textcolor{BurntOrange}{\textbf{A:}} Yes. Sources differ in spatial resolution, spectral response, acquisition conditions, and modality availability. Captions combine existing annotations with model-generated descriptions based primarily on RGB visualizations; these descriptions may omit non-visible spectral or radar-specific information. 
	
	\item \textit{``Are there tasks for which the dataset should not be used?''} 
	
	\textcolor{BurntOrange}{\textbf{A:}} RSI-MM is intended for research on remote sensing representation and generation. Synthetic outputs should not be presented as actual sensor measurements or used as sole evidence for conclusions that require verified observations. 
\end{enumerate}

\subsection{Distribution}
Dataset creators should provide answers to these questions prior to distributing the dataset either internally within the entity on behalf of which the dataset was created or externally to third parties.
\begin{enumerate}
	\item \textit{``Will the dataset be distributed to third parties outside of the entity (\textit{e.g.}, company, institution, organization) on behalf of which the dataset was created?''} 
	
	\textcolor{BurntOrange}{\textbf{A:}} The benchmark will be made publicly accessible to the research community, subject to the licenses and terms of the underlying source datasets.
	
	\item \textit{``How will the dataset be distributed (\textit{e.g.}, tarball on website, API, GitHub)?''} 
	
	\textcolor{BurntOrange}{\textbf{A:}} We will provide RSI-MM in the GitHub and the Hugging face repository.
	
	\item \textit{``When will the dataset be distributed?''} 
	
	\textcolor{BurntOrange}{\textbf{A:}} We will create a repository to release the data once the paper is officially published.
	
	\item \textit{``Will the dataset be distributed under a copyright or other intellectual property (IP) license, and/or under applicable terms of use (ToU)?''} 
	
	\textcolor{BurntOrange}{\textbf{A:}} Yes. The newly created RSI-MM instructions, annotations, and metadata will be released under the Creative Commons Attribution-NonCommercial-ShareAlike 4.0 International License, while source imagery remains subject to the licenses and terms of the original datasets.
	
	\item \textit{``Have any third parties imposed IP-based or other restrictions on the data associated with the instances?''} 
	
	\textcolor{BurntOrange}{\textbf{A:}} Yes. The underlying source images are subject to the licenses and terms of their original datasets. RSI-MM will preserve source attribution and document applicable usage constraints in the release.
	
	\item \textit{``Do any export controls or other regulatory restrictions apply to the dataset or to individual instances?''} 
	
	\textcolor{BurntOrange}{\textbf{A:}} We are not aware of export-control restrictions specific to RSI-MM beyond the licenses and terms of the underlying source datasets. Users are responsible for ensuring compliance with applicable local regulations.    
\end{enumerate}
 
\subsection{Maintenance}
As with the questions in the previous section, dataset creators should provide answers to these questions prior to distributing the dataset. The questions in this section are intended to encourage dataset creators to plan for dataset maintenance and communicate this plan to dataset consumers.
\begin{enumerate}
	\item \textit{``Who will be supporting/hosting/maintaining the dataset?''} 
	
	\textcolor{BurntOrange}{\textbf{A:}} The authors of this work will support, host, and maintain the dataset.
	
	\item \textit{``How can the owner/curator/manager of the dataset be contacted (\textit{e.g.}, email address)?''} 
	
	\textcolor{BurntOrange}{\textbf{A:}} They can be contacted via the email addresses listed on the paper or webpage.
	
	\item \textit{``Is there an erratum?''} 
	
	\textcolor{BurntOrange}{\textbf{A:}} There is no erratum for the initial release; updates and known issues will be documented on the project repository.
	
	\item \textit{``Will the dataset be updated (\textit{e.g.}, to correct labeling errors, add new instances, delete instances)?''} 
	
	\textcolor{BurntOrange}{\textbf{A:}} Future updates (if any) will be posted on the dataset website.
	
	\item \textit{``Will older versions of the dataset continue to be supported/hosted/maintained?''} 
	
	\textcolor{BurntOrange}{\textbf{A:}} 
	
	Yes. Older versions will be archived when feasible, and updates will be documented on the project repository.
	
	\item \textit{``If others want to extend/augment/build on/contribute to the dataset, is there a mechanism for them to do so?''} 
	
	\textcolor{BurntOrange}{\textbf{A:}} Yes, we will provide detailed instructions for future extensions.
\end{enumerate}

\setcounter{equation}{0}
\renewcommand{\theequation}{\thesection.\arabic{equation}} 
\renewcommand{\thesection}{G}
\section{Discussions, Limitations, and Future Work}\label{sec:G} 
\noindent{\textbf{Limitations and broader impact.}} OmniRSI aims to unify multimodal remote sensing reconstruction, generation, and translation, but several limitations remain. First, partial modality pairing limits supervision for certain cross-modal mappings. Second, modality-aware slots provide a consistent representation structure but do not explicitly harmonize sensor-specific spectral responses, spatial resolutions, or radiometric characteristics. Consequently, generated observations in missing slots should be interpreted as conditional predictions rather than physically calibrated measurements. Third, captions derived from RGB visualizations may overlook spectral or radar-specific information and introduce annotation errors. Last, full backbone fine-tuning requires substantial computational resources, while CLA improves the performance but requires two velocity evaluations per sampling step, increasing inference cost. Despite these limitations, OmniRSI provides a foundation for learning shared generative priors from heterogeneous and incomplete observations. Its ability to synthesize missing modalities may support data augmentation and Earth observation research.

\noindent{\textbf{Future Work.}} Future research could pursue several directions. (1) Expanding and balancing RSI-MM with more geographically diverse, temporally aligned multimodal observations and improving caption quality through modality-aware annotation and verification. (2) Incorporating sensor metadata, spectral response functions, and physical constraints to improve cross-sensor consistency and the reliability of generated spectral information. (3) Developing more efficient architectures and sampling strategies, including distillation of cyclic latent alignment, to reduce training and inference costs and support higher-resolution imagery. (4) Introducing uncertainty estimation for modality combinations with limited or no paired supervision.

\end{document}

%% file: math_commands.tex
\usepackage{amsmath,amsfonts,bm}

\def\eqref#1{equation~\ref{#1}}

\def\1{\bm{1}}

\DeclareMathAlphabet{\mathsfit}{\encodingdefault}{\sfdefault}{m}{sl}
\SetMathAlphabet{\mathsfit}{bold}{\encodingdefault}{\sfdefault}{bx}{n}

